\documentclass[journal]{IEEEtran}
\usepackage{amsmath,amsfonts}
\usepackage{algorithmic}
\usepackage{array}
\usepackage{amsmath}
\usepackage{amssymb}
\usepackage{mathrsfs}
\usepackage[caption=false,font=normalsize,labelfont=sf,textfont=sf]{subfig}
\usepackage{textcomp}
\usepackage{stfloats}
\usepackage{multirow}
\usepackage{booktabs}
\usepackage{cite}
\usepackage{color}
\usepackage{makecell}
\usepackage{url}
\usepackage{ulem}
\usepackage{verbatim}
\usepackage{graphicx}
\newcommand{\verticaltext}[1]{%
    \rotatebox[origin=c]{90}{#1}%
}
\def\BibTeX{{\rm B\kern-.05em{\sc i\kern-.025em b}\kern-.08em
    T\kern-.1667em\lower.7ex\hbox{E}\kern-.125emX}}
\usepackage{balance}
\begin{document}
\title{Power Line Aerial Image Restoration under Adverse Weather: Datasets and Baselines}
\author{ Sai Yang, Bin Hu, Bojun Zhou, Fan Liu,~\IEEEmembership{Member,~IEEE,}  Xiaoxin Wu, Xinsong Zhang~\IEEEmembership{Member,~IEEE,} Juping Gu,~\IEEEmembership{Senior Member,~IEEE,} Jun Zhou,~\IEEEmembership{Senior Member,~IEEE} 
        % <-this % stops a space
    \thanks{This work was partially supported by the National Nature Science Foundation of China (62372155, 62273188, U2066203, 61973178, 52377117), Nantong Science and Technology Program Project (MS2023060)}
    \thanks{ Corresponding author: Fan Liu and Juping Gu}
	\thanks{Sai Yang, Xiaoxin Wu, Xinsong Zhang are with the School of Electrical Engineering and Automation, Nantong University, Nantong, 226019, China (e-mail: yangsai@ntu.edu.cn, wu.xx@ntu.edu.cn, zhang.xs@ntu.edu.cn)}
  \thanks{Bin Hu is with the School of Artificial Intelligence and Computer Science, Nantong University, Nantong, 226019, China (e-mail: hubin@ntu.edu.cn)}
  \thanks{Bojun Zhou and Juping Gu are the School of Information Science and Technology, Nantong University, Nantong, 226019, China (e-mail: zhoubj@ntu.edu.cn, gu.jp@ntu.edu.cn)}
   \thanks{Fan Liu is with the College of Computer and Software Engineering, Hohai University, Nanjing 210098, China (e-mail: fanliu@hhu.edu.cn).}
 \thanks{Juping Gu is with the School of Electronic and Information Engineering, Suzhou University of Science and Technology, Suzhou, 215009, China, (e-mail: gu.jp@ntu.edu.cn).}
 \thanks{Jun Zhou is with the School of Information and Communication Technology, Griffith University, Nathan, QLD 4111, Australia (email:jun.zhou@griffith.edu.au).}
}

\markboth{Journal of \LaTeX\ Class Files,~Vol.~18, No.~9, September~2020}%
{How to Use the IEEEtran \LaTeX \ Templates}

\maketitle

\begin{abstract}
 Power Line Autonomous Inspection (PLAI) plays a crucial role in the construction of smart grids due to its great advantages of low cost, high efficiency, and safe operation. PLAI is completed by accurately detecting the electrical components and defects in the aerial images captured by Unmanned Aerial Vehicles (UAVs). However, the visible quality of aerial images is inevitably degraded by adverse weather like haze, rain, or snow, which are found to drastically decrease the detection accuracy in our research. To circumvent this problem, we propose a new task of Power Line Aerial Image Restoration under Adverse Weather (PLAIR-AW), which aims to recover clean and high-quality images from degraded images with bad weather thus improving detection performance for PLAI. In this context, we are the first to release numerous corresponding datasets, namely, HazeCPLID, HazeTTPLA, HazeInsPLAD for power line aerial image dehazing, RainCPLID, RainTTPLA, RainInsPLAD for power line aerial image deraining, SnowCPLID, SnowInsPLAD for power line aerial image desnowing, which are synthesized upon the public power line aerial image datasets of CPLID, TTPLA, InsPLAD following the mathematical models. Meanwhile, we select numerous state-of-the-art methods from image restoration community as the baseline methods for PLAIR-AW. At last, we conduct large-scale empirical experiments to evaluate the performance of baseline methods on the proposed datasets. The proposed datasets and trained models are available at \textcolor[rgb]{1,0.1,0.5}{https://github.com/ntuhubin/PLAIR-AW}.
\end{abstract}


\section{Introduction}
\begin{figure}[h!]
		\includegraphics[width=1\linewidth]{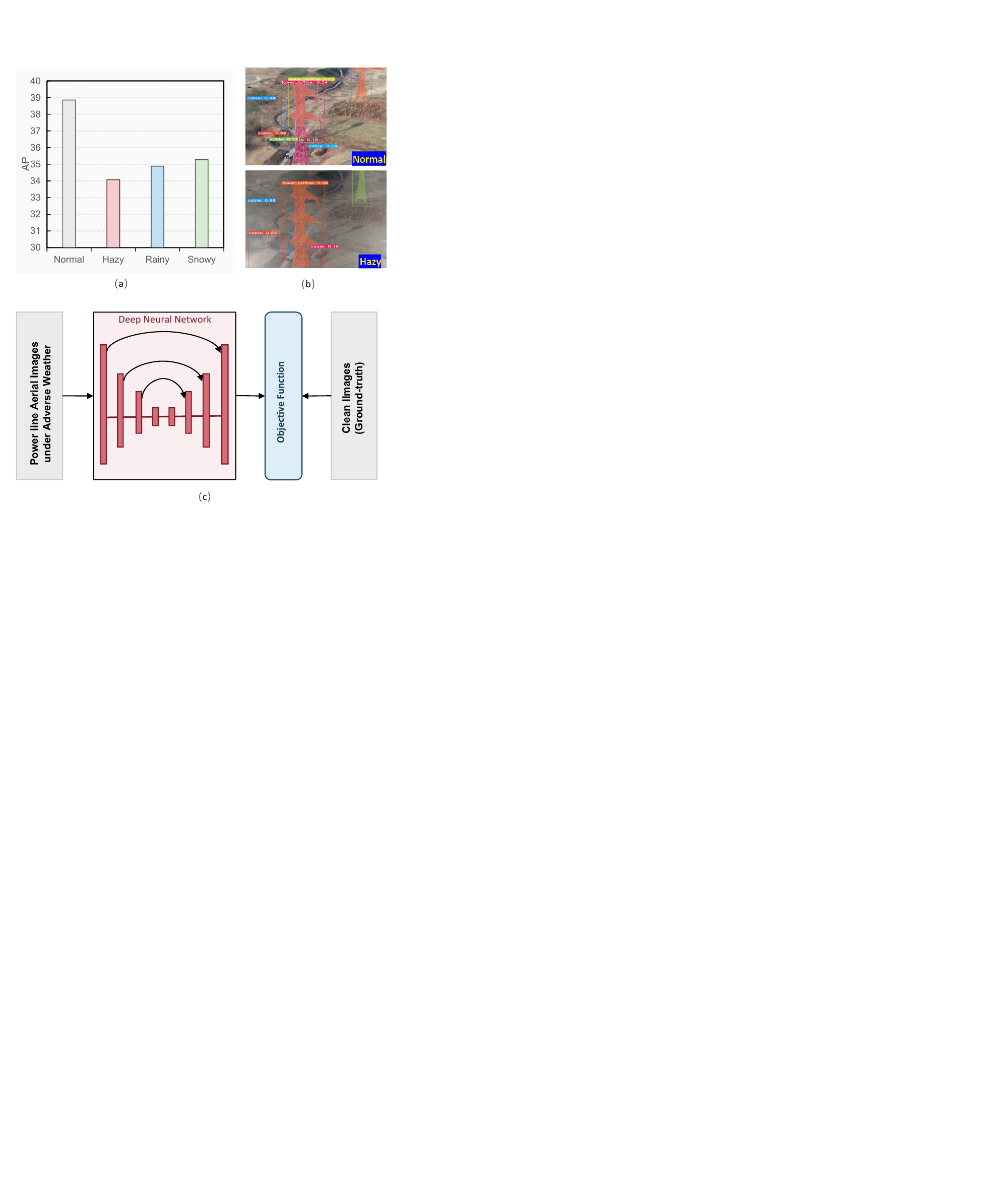}
		\caption{(a) Average precision (AP) of the advanced real-time instance segmentation model on a popular dataset of TTPLA under Normal, Hazy, Rainy, and Snowy conditions, respectively. There is a substantial decline in adverse weather compared with the normal situation. (b)~The visual comparison between normal and hazy conditions, suggests that missed and false detection exists in hazy cases. (c) The general solution framework is based on deep learning for Power Line Aerial Image Restoration under Adverse Weather (PLAIR-AW). The power line aerial images under adverse weather are input into the deep neural network, which outputs the restored images. Then, the objective function is established between the ground-truths and the restored images. The deep neural network is optimized with the objective function. After training, the deep neural network is deployed to do the PLAIR-AW test. }
		\label{Fig1}
	%\end{center}
    \end{figure}
\IEEEPARstart{U}{ndoubtedly}, smart grid has become the common development trend for future power systems around the world~\cite{zhu2023fllf,hu2022reinforcement}. The overhead power transmission lines are critical infrastructures of the power system, and their working status directly affects the stability and reliability of the entire power system. Power line inspection can timely identify and eliminate potential hazards thus avoiding unplanned outages. The traditional way of power line inspection is the manual inspection, which suffers from the disadvantages of low efficiency and high risk. Instead, Power Line Autonomous Inspection (PLAI) with Unmanned Aerial Vehicles (UAVs) has gradually become the mainstream inspection way, thanks to the low cost, high efficiency, and safe operation. PLAI is of great importance to guarantee the safe, reliable, and efficient operation of the smart grid, which has received increasing attention from the community~\cite{hosseini2020intelligent,Wei2023Detecting}.

With the developments of UAVs and Artificial Intelligence (AI), PLAI is generally implemented by detecting key electrical components and defects with deep learning techniques in aerial images captured by UAVs~\cite{cao2023few,li2023robust,yu2022object,yang2021asymmetric,liu2023transcending}. Since adverse weather such as haze, rain, and snow are common phenomena, the aerial images captured under such conditions are inevitably subject to severe visibility degradation of color fidelity, blurring, low contrast, and obscured objects, thereby seriously
reducing the detection accuracy. To support the above argument, we conducted a pilot study on the popular power line inspection dataset of TTPLA~\cite{Abdelfattah2020Ttpla} with the state-of-the-art real-time instance segmentation model of YOLACT~\cite{bolya2019yolact}. This experiment was implemented with normal images, hazy images, rain images, and snowy images, respectively. We select the Average Precision (AP) as the performance metric and report the results in Fig.~\ref{Fig1} (a), where AP in hazy, rainy, and snowy conditions are lower than the normal case. The visual comparison between the normal condition and the hazy condition is illustrated in Fig.~\ref{Fig1} (b), revealing that there appeared missed and false detection in the hazy case. 
    
To improve the detection performance for PLAI, we propose a new task of Power Line Aerial Image Restoration under Adverse Weather (PLAIR-AW), which attempts to recover clean and high-quality images from degraded images captured by UAVs in adverse weather conditions. PAIR-AW is a highly challenging task because of its ill-posed nature. Considering the powerful nonlinear representation capabilities of deep neural networks, we formulate the general solution framework based on deep learning for PAIR-AW in Fig.~\ref{Fig1} (c). It follows that the three key factors are the data, model architecture, and objective function. Wherein, the data is the prerequisite of the solution, while model architecture and objective function are determined by the specific method. Therefore, we provide the corresponding datasets and baselines for the in-depth research for PAIR-AW. Specifically, according to weather type, PAIR-AW is further subdivided into three sub-tasks, i.e. power line aerial image dehazing, power line aerial image draining, and power line aerial image desnowing. We establish datasets for each separate sub-task based on the public power line aerial image datasets of  CPLID~\cite{Tao2020Detection}, TTPLA~\cite{Abdelfattah2020Ttpla}, InsPLAD~\cite{Silva2023OInsPLAD}. Following the Atmospheric Scattering Model (ASM)~\cite{narasimhan2003contrast,nayar1999vision}, we construct synthetic datasets of HazeCPLID, HazeTTPLA, and HazeInsPLAD for the dehazing task. Following the Comprehensive Rain Model (CRM)~\cite{yang2019frame}, we construct synthetic datasets of RainCPLID, RainTTPLA, and RainInsPLAD for deraining tasks. Following the mathematical model proposed by Liu et al.~\cite{2017DesnowNet}, we construct synthetic datasets of SnowCPLID, and SnowTTPLA for the desnowing task. Otherwise, we select numerous state-of-the-art methods in the image restoration community as the baseline methods for PLAIR-AW. At last, we conduct large-scale empirical experiments to evaluate the performance of baseline methods on the proposed datasets. 

In summary, our main contributions are as follows:
\begin{itemize}
\item This is the first to propose the new task of Power Line Aerial Image Restoration under Adverse Weather (PLAIR-AW), which attempts to recover clean and high-quality images from degraded images captured by UAVs in adverse weather conditions. This research is of great importance to meet the realistic demand for power line autonomous inspection.
\item We are the first to generate numerous power line aerial image datasets under multiple adverse weathers, which can provide strong support for the research of PLAIR-AW. Meanwhile, we also provide numerous excellent baseline methods for the new PLAIR-AW task.
\item We conduct large-scale empirical experiments to evaluate the performance of
baseline methods on the proposed datasets in both single-one and all-in-one settings.  
\end{itemize}

The remainder of this paper is organized as follows.  In section $\rm\uppercase\expandafter{\romannumeral 2}$, we mainly formulate the generation process of datasets for PLAIR-AW task. In Section $\rm\uppercase\expandafter{\romannumeral 3}$, we detail the baseline methods for PLAIR-AW task. In Section $\rm\uppercase\expandafter{\romannumeral 4}$, we describe the extensive experimental comparison results among the baseline methods. Finally, we present a conclusion with some future research directions in Section $\rm\uppercase\expandafter{\romannumeral 5}$.

\begin{table*}[ht]
    \renewcommand\tabcolsep{1.5pt} 
    \caption{Summarization of the proposed public datasets for power line aerial image restoration under adverse weather (PLAIR-AW) task. '-' means there is no subset in this dataset.}
    \label{table1}
    \begin{center}
    \begin{tabular}{p{1.5cm}cccccc}
	\hline
	% \multirow{2}{*}{Name}，2为所占的行数，此语句可以使得内容垂直居中
	% \multicolumn{2}{c|}{Flag}，2为所占的列数，格式由第二个{}控制
	% \cline{2-3}指本行的2,3列画横线

    Dataset&Type&Subset&Size&Numbers&train/test& Download Link \\
    \hline
    HazeCPLID&Dehazing&-&84×84&848&700/148&https://github.com/ntuhubin/PLAIR-AW/blob/main/Dehazing-DataSet.md\\
    HazeTTPLA&Dehazing&-&512×512&1242&1000/242&https://github.com/ntuhubin/PLAIR-AW/blob/main/Dehazing-DataSet.md\\
   HazeInsPLAD&Dehazing&-&1920×1080&10,607&7981/2626&https://github.com/ntuhubin/PLAIR-AW/blob/main/Dehazing-DataSet.md\\
   RainCPLID&Deraining&RainCPLID-L/H&84×84&848&700/148&https://github.com/ntuhubin/PLAIR-AW/blob/main/Deraining-Datasets.md\\
    RainTTPLA&Deraining&RainTTPLA-L/H&512×512&1242&1000/242&https://github.com/ntuhubin/PLAIR-AW/blob/main/Deraining-Datasets.md \\
    RainInsPLAD&Deraining&RainInsPLAD-L/H&1920×1080&10,607&7981/2626&https://github.com/ntuhubin/PLAIR-AW/blob/main/Deraining-Datasets.md\\
    SnowCPLID&Desnowing&SnowCPLID-S/M/L&84×84&848&700/148&https://github.com/ntuhubin/PLAIR-AW/blob/main/Desnowing-Datasets.md\\
    SnowTTPLA&Desnowing&SnowTTPLA-S/M/L&512×512&1242&1000/242&https://github.com/ntuhubin/PLAIR-AW/blob/main/Desnowing-Datasets.md\\
	\hline
    \end{tabular}
    \end{center}
\end{table*}
 \begin{figure*}[t]
	%\begin{}
	%\vspace{-0.5cm}
		%\fbox{\rule{0pt}{2in} \rule{.9\linewidth}{0pt}}
		%\includegraphics[scale=0.4]{fig2.jpg}
		\includegraphics[width=1\linewidth]{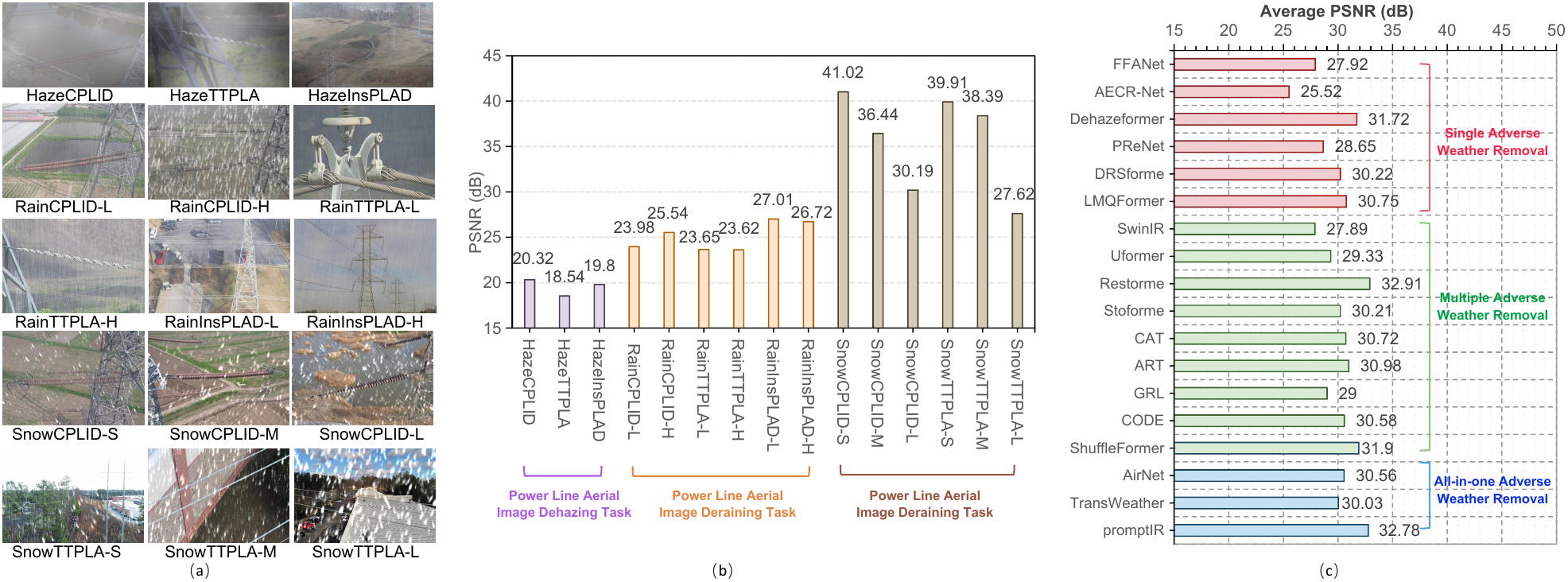}
		%	\vspace*{-0.5	\baselineskip}
		\caption{ (a) The exemplar images of the proposed HazeCPLID, HazeTTPLA, HazeInsPLAD, RainCPLID-L, RainTTPLA-H, RainTTPLA-L, RainTTPLA-H, RainInsPLAD-L, RainInsPLAD-H, SnowCPLID-S, SnowCPLID-M, SnowCPLID-H, SnowTTPLA-S, SnowCPLID-M, SnowCPLID-H datasets. (b) The information loss of each proposed dataset is compared with its clean counterpart, which is measured by the Peak Signal-to-Noise Ratio (PSNR). (c) The performance ranking of the baseline methods.}
		\label{Fig2}
	%\end{center}
\end{figure*}
\begin{figure*}[htbp]
	%\begin{}
	%\vspace{-0.5cm}
		%\fbox{\rule{0pt}{2in} \rule{.9\linewidth}{0pt}}
		%\includegraphics[scale=0.4]{fig2.jpg}
		\includegraphics[width=1\linewidth]{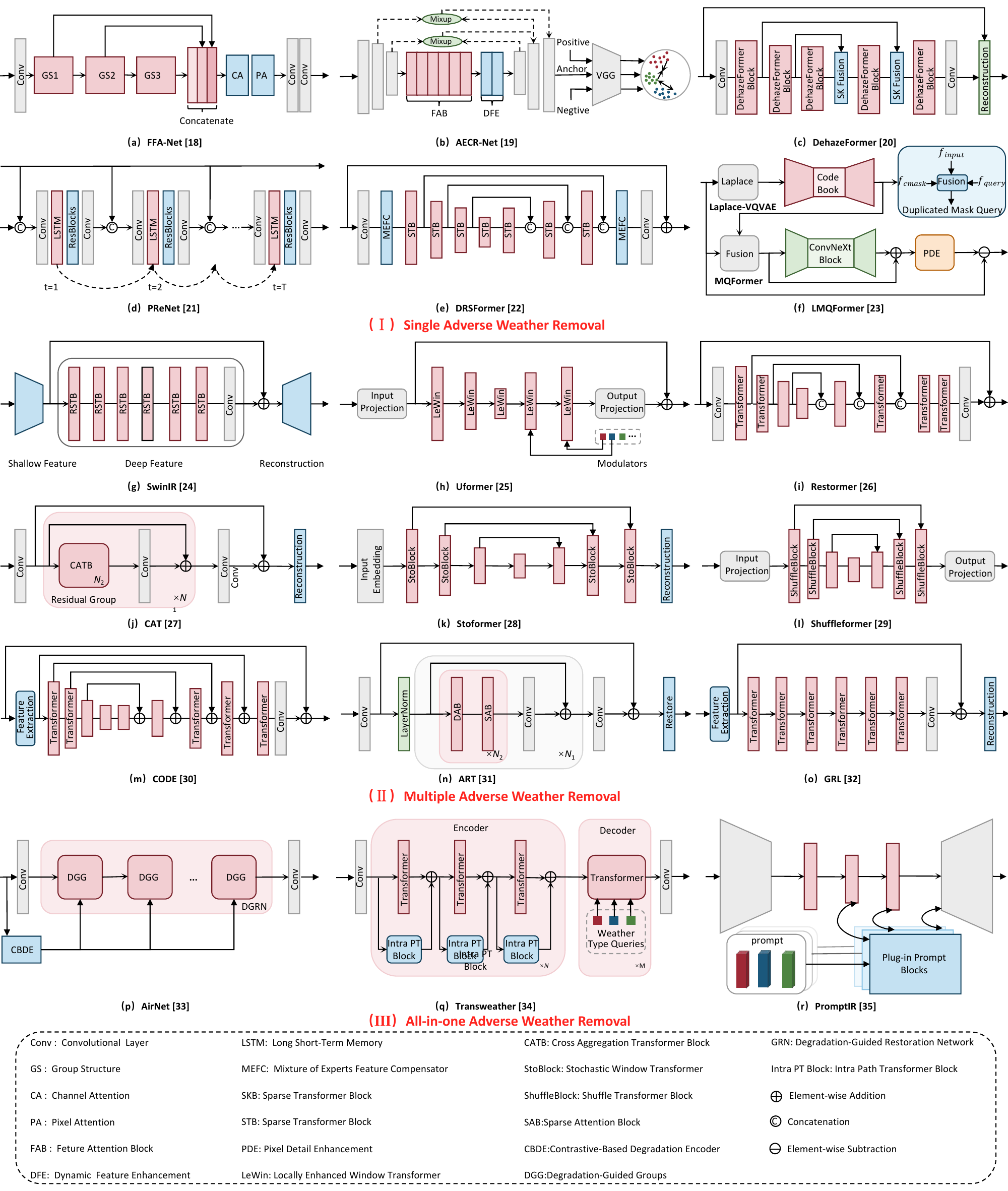}
		%	\vspace*{-0.5	\baselineskip}
		\caption{Illustration of baseline methods for the new Power Line Aerial Image Restoration under Adverse Weather (PLAIR-AW) task. These baseline methods are broadly categorized into three families of ($\rm\uppercase\expandafter{\romannumeral 1}$) Single adverse weather removal, ($\rm\uppercase\expandafter{\romannumeral 2}$) Multiple adverse weather removal, and ($\rm\uppercase\expandafter{\romannumeral 3}$) All-in-one adverse weather removal. In the first family, we illustrate the representative methods for each specific adverse weather removal, namely, (a) FFANet~\cite{qin2020ffa}, (b) AECR-Net~\cite{wu2021contrastive}, (c) Dehazeformer~\cite{song2023vision} for power line aerial image dehazing, (d) PReNet~\cite{ren2019progressive}, (e) DRSformer~\cite{chen2023learning} for power line aerial image deraining, (f) LMQFormer~\cite{lin2023lmqformer} for power line aerial image desnowing. The representative methods in the second family include (g) SwinIR~\cite{liang2021swinir}, (h) Uformer~\cite{wang2022uformer}, (i) Restormer~\cite{zamir2022restormer} (j) CAT~\cite{chen2022cross}, (k) Stoformer~\cite{xiao2022stochastic}, (l) ShuffleFormer~\cite{xiao2023random}, (m) CODE~\cite{zhao2023comprehensive}, (n) ART~\cite{2023ART}, (o) GRL~\cite{li2023efficient}. The representative methods in the third family include (p) AirNet~\cite{li2022all}, (q) TransWeather~\cite{valanarasu2022transweather}, (r) PromptIR~\cite{potlapalli2023promptir}. }
		\label{Fig3}
	%\end{center}
\end{figure*}

\begin{table*}[ht]
    \renewcommand\tabcolsep{3.6pt} 
    \caption{Summarization of The baseline methods. ``Con", ``SSIM", ``Char", ``Per", ``Edge", and ``SC" stand for contrastive loss,  structural similarity loss, charbonnier loss, perceptual loss, edge loss, and supervised contrastive loss, respectively.}
     \scriptsize
    \label{table2}
    \begin{center}
    \begin{tabular}{p{0.5cm}||llllllll}
	\hline
	% \multirow{2}{*}{Name}，2为所占的行数，此语句可以使得内容垂直居中
	% \multicolumn{2}{c|}{Flag}，2为所占的列数，格式由第二个{}控制
	% \cline{2-3}指本行的2,3列画横线

    \textbf{Type}&\textbf{Method}&\textbf{Main Blocks}&\textbf{Structure}&\textbf{Loss Function}& \textbf{Download Link}&\textbf{Venu}&\textbf{Year} \\
    \hline
    \multirow{6}{*}{\verticaltext{SAWR}}&FFANet~\cite{qin2020ffa}&Attention&Asymmetric&$\mathscr{L}_1$&https://github.com/zhilin007/FFA-Net& AAAI&2020\\
    &AECR-Net~\cite{wu2021contrastive}&Attention&U-Net&$\mathscr{L}_1+\mathscr{L}_{Con}$&https://github.
com/GlassyWu/AECR-Net&CVPR&2021\\
&DehazeFormer~\cite{song2023vision}&Transformer&U-Net&$\mathscr{L}_1+\mathscr{L}_{SSIM}+\mathscr{L}_{Per}+\mathscr{L}_{Adv}$&https://github.com/IDKiro/DehazeFormer&TIP&2023\\
&PReNet~\cite{ren2019progressive}&Convolutional&Multi-stage&$\mathscr{L}_2+\mathscr{L}_{SSIM}$&https://github.com/csdwren/PReNet&CVPR&2019\\
&DRSformer~\cite{chen2023learning}&Transformer&U-Net&$\mathscr{L}_1$&https://github.
com/cschenxiang/DRSformer&CVPR&2023\\
&LMQFormer~\cite{lin2023lmqformer}&Transformer&Pysic-aware&$\mathscr{L}_{Char}+\mathscr{L}_{Per}+\mathscr{L}_{Edge}$&https://github.com/StephenLinn/LMQFormer&TIP&2023\\
	\hline
 \multirow{9}{*}{\verticaltext{MAWR}}&SwinIR~\cite{liang2021swinir}&Transformer&Asymmetric&$\mathscr{L}_{Char}$&https://github.com/JingyunLiang/SwinIR&ICCVW&2021\\
&Uformer~\cite{wang2022uformer}&Transformer&U-Net&$\mathscr{L}_{Char}$&https:
//github.com/ZhendongWang6/Uformer&CVPR&2022\\
&Restormer~\cite{zamir2022restormer}&Transformer&U-Net&$\mathscr{L}_{1}$&https://github.com/swz30/Restormer&CVPR&2022\\
&CAT~\cite{chen2022cross}&Transformer&U-Net&$\mathscr{L}_{Char}$& https://github.com/zhengchen1999/CAT&NIPS&2022\\
&Stoformer~\cite{xiao2022stochastic}&Transformer&U-Net&$\mathscr{L}_{Char}$&https://github.com/jiexiaou/Stoformer&NIPS&2022\\
&ShuffleFormer~\cite{xiao2023random}&Transformer&U-Net&$\mathscr{L}_{Char}$&https://github.com/jiexiaou/
ShuffleFormer&ICML&2023\\
&CODE~\cite{zhao2023comprehensive}&Transformer&U-Net&$\mathscr{L}_{2}$&https://github.com/XLearning-SCU/2023-CVPR-CODE&CVPR&2023\\
&ART~\cite{2023ART}&Transformer&Asymmetric&$\mathscr{L}_{1}+\mathscr{L}_{Char}$&https://github.com/gladzhang/ART&ICLR&2023\\
&GRL~\cite{li2023efficient}&Transformer&U-Net&$\mathscr{L}_{1}$&https://github.com/ofsoundof/GRL-Imag Restoration.git&CVPR&2023\\
\hline 
\multirow{3}{*}{\verticaltext{AAWR}}
&AirNet~\cite{li2022all}&Convolutional&Asymmetric&$\mathscr{L}_{1}+\mathscr{L}_{SC}$&https://github.
com/XLearning-SCU/2022-CVPR-AirNet&CVPR&2022\\
&TransWeather~\cite{valanarasu2022transweather}&Transformer&Asymmetric&$\mathscr{L}_1 + \mathscr{L}_{Per}$&https://github.com/jeya-maria-jose/TransWeather&CVPR&2022\\
&PromptIR~\cite{potlapalli2023promptir}&Transformer&U-Net&$\mathscr{L}_1$&https://github.com/va1shn9v/PromptIR&NIPS&2023\\
\hline
    \end{tabular}
    \end{center}
\end{table*}
\section{Problem Setting and Datasets}
As mentioned in the Introduction, sufficient paired degraded aerial images and clean images are prerequisites for deep-learning-based methods. However,
it is impossible to capture the same scene by UAVs under normal and adverse weather conditions at the same time. Alternatively, we synthesize the hazy, rainy, and snowy images upon clean ones with simulated components in mathematical models to possibly approach the real adverse weather environment. The details are as follows:
\subsection{Power Line Aerial Image Dehazing}
\subsubsection{Mathematical Models}
When meeting the hazy weather, the sunlight toward the camera will be changed to be atmosphere light $A$ because of the floating particles. Meanwhile, light from the scene is attenuated to be medium transmission map $T$. The hazy images are formed by the joint action of the atmosphere light $A$ and medium transmission map $T$, which can be well described by the Atmospheric Scattering Model (ASM) with the following formula: 
\begin{equation}
    \label{eq:equ1}
   H(x)= I(x)T(x)+A(1-T(x)),
\end{equation}
where, $x$ is the pixel location, $H(x)$ and $I(x)$ represent the hazy image and its corresponding clean image, respectively. $A$ is the global atmosphere light. $T(x)$ is the medium transmission map, which is mainly determined by the scene depth of the image with the following formula:
\begin{equation}
    \label{eq:equ2}
   T(x)=e^{-\beta d(x)},
\end{equation}
where, $\beta$ is the atmosphere scattering parameter, and $d(x)$ is the scene depth. 
\subsubsection{Power Line Aerial Image Dehazing Datasets}
According to Equation~(\ref{eq:equ1}) and Equation~(\ref{eq:equ2}), we can see that the hazy process is determined by three parameters of $A$, $\beta$ and $d(x)$. Therefore, to simulate various scenarios of hazy conditions,  the values of $A$ are randomly set from 0.4 to 0.6 and the values of $\beta$ are randomly set from $5e^{-6}$ to $7e^{-6}$. Also, $d(x)$ is selected as the average distance of the UAVs from the ground. The above operations are implemented on the public CPLID~\cite{Tao2020Detection}, TTPLA~\cite{Abdelfattah2020Ttpla}, InsPLAD~\cite{Silva2023OInsPLAD} datasets, thus producing the following dehazing datasets:
\begin{itemize}
\item [*] HazeCPLID: This dataset is created based on CPLID~\cite{Tao2020Detection}, which consists of 848 pairs of hazy and clean images. The size of images is $84 \times 84$. The total images are divided into 700 and 148 for training and testing, respectively. 

\item [*]HazeTTPLA: This dataset is created based on TTPLA~\cite{Abdelfattah2020Ttpla}, which consists of 1242 paired hazy and clean images with the size of 512×512 pixels. The total images are divided into 1000 and 242 for training and testing, respectively. 

\item [*] HazeInsPLAD: This dataset is created based on InsPLAD~\cite{Silva2023OInsPLAD}, which consists of 10,607 paired snowy and clean images in 1920×1080 resolution. The total images are divided into 7981 and 2626 images for training and testing.
\end{itemize}
\subsection{Power Line Aerial Image Deraining}
\subsubsection{Mathematical Models}
Owing to complication of the rainy condition, various mathematical models, such as Additive Composite Model (ACM)~\cite{2020Single,li2016rain}, Screen-Blend Model (SBM)~\cite{luo2015removing}, Occlusion-aware Hybrid Rain Model (OHRM)~\cite{liu2018rain}, Comprehensive Rain Model (CRM)~\cite{yang2019frame}, have been proposed. Since ACM, SBM, OHRM can be viewed as a simplified case of CRM, we use CRM as the mathematical model to describe the physical process of rainy images:
\begin{equation}
    \label{eq:equ3}
    \begin{aligned}
  R(x)=(1- \alpha(x))[\beta I(x)+(1-\beta)A+&\sum_{l=1}^{L}S_l(x)]+\\
  &\alpha(x)M(x),
  \end{aligned}
\end{equation}
where $R(x)$ and $I(x)$ denote the rainy image and its corresponding clean image, respectively. $S_l(x)$ is the $l$-th rain streak layer. $\beta$ 
and $A$ respectively denote atmospheric transmission and the global atmospheric light. $M(x)$ is the rain reliance map and $\alpha(x)$ is an alpha matting map, which is:
\begin{equation}
    \label{eq:equ4}
    \begin{aligned}
     \alpha(x)=\begin{cases}
   1, \quad \quad if \quad  x\in \Omega_s, \\
   0,\quad \quad if \quad x\notin \Omega_s,
   \end{cases}.
   \end{aligned}
\end{equation}
where $\Omega_s$ is defined as the rain-occluded region.
\subsubsection{Power Line Aerial Image Deraining Datasets}
From Equation~(\ref{eq:equ3}), it notes that CRM thoroughly comprises the key factors in the complicated atmospheric process of the rain, i.e., rain streaks, raindrops, and mist-like phenomenon. Consequently, we synthesize sharp lines and transparent circles to simulate the rain streaks and raindrops by using Photoshop. The combination of the rain streaks and raindrops is referred to as the rain mask. Also, we adopt the synthesized procedure in the dehazing dataset to create the mist-like phenomenon. The synthesized rain mask and haze are superposed upon the clean images from the public CPLID~\cite{Tao2020Detection}, TTPLA~\cite{Abdelfattah2020Ttpla}, InsPLAD~\cite{Silva2023OInsPLAD} datasets, we get the following deraining datasets:
\begin{itemize}
\item [*] RainCPLID: This dataset is generated upon CPLID, which has two subsets of RainCPLID-L and RainCPLID-H. The former is produced with light rain streaks, while the latter is produced with heavy rain streaks. Both subsets include 848 pairs of rainy and clean images with the size of $84 \times 84$. The images are split into 700 for training, and 148 for testing. 
\item [*] RainTTPLA: This dataset is generated upon TTPLA, having two subsets of RainTTPLA-L and RainTTPLA-H. The former is produced with light rain streaks, while the latter is produced with heavy rain streaks. Both subsets contain 1242 pairs of rainy and clean images with a resolution of 512×512. The total images are divided into 1000, and 242 for training and testing.
\item [*] RainInsPLAD: This dataset is generated upon InsPLAD, which has two subsets of RainInsPLAD-L and RainInsPLAD-H. The former is produced with light rain streaks, while the latter is produced with heavy rain streaks. Both subsets contain 10,607 pairs of rainy and clean images in 1920×1080 resolution. The total images are divided into 7981 and 2626 images for training and testing, respectively. 
\end{itemize}
\subsection{Power Line Aerial Image Desnowing}
\subsubsection{Mathematical Models}
In snowy conditions, Liu et al.~\cite{2017DesnowNet} established the following mathematical model:
\begin{equation}
    \label{eq:equ5}
    S(x)=J(x)(1-Z(x)) + C(x)Z(x),
\end{equation}
where, $S(X)$ and $J(x)$ denote the snowy image and its corresponding clean image. $C(x)$ represents the snow flakes and $Z(x)$ is a binary mask indicating the location of snow.
\subsubsection{Power Line Aerial Image Desnowing Datasets}
Following Equation~(\ref{eq:equ5}), we simulate the snow mask and aberration map to generate the snowy images. Specifically, we use Photoshop to generate snowflakes and streaks with different transparencies and sizes in different locations, and then adopt Gaussian blurring on snow particles to produce the aberration map. According to the density of snow particles, the snow mask is further divided into three types small, medium, and large. With the snow mask, the desnowing datasets upon the public CPLID~\cite{Tao2020Detection} and TTPLA~\cite{Abdelfattah2020Ttpla} are produced as follows:
\begin{itemize}
\item [*]SnowCPLID: This dataset is constructed based on CPLID, which provides three kinds of snowy images with different sizes of snow particles, namely, SnowCPLID-S, SnowCPLID-M, and SnowCPLID-L. Each subset has a total number of 848 pairs of snowy and clean images, which are divided into 700, and 148 for training and testing, respectively. The size of images in training and testing sets is 84 × 84.

\item [*] SnowTTPLA: This dataset is constructed based on TTPLA, which provides three kinds of snowy images with different sizes of snow particles, namely, SnowTTPLA-S, SnowTTPLA-M, and SnowTTPLA-L. Each subset consists of 1,242 paired snowy and clean images totally, which are partitioned into 1,000 and 242 for training and testing, respectively. The images in training and testing sets are in size of 512×512.
\end{itemize}
\textbf{Summary and Challenges:} The details about the task type, subset, image size, image numbers, splits protocol as well as the download link of all the proposed datasets are summarized in Table~\ref{table1}. The exemplar images of the proposed datasets are shown in Fig.~\ref{Fig2} (a). Moreover, we present the characteristics of the proposed datasets in the following aspects: (1) In HazeCPLID, RainCPLID, and SnowCPLID datasets, the main foreground objects in the images are insulators, which are further categorized into two types of normal and missing-cap fault. The background of the images covers the scenes of cities, rivers, fields, and mountains. In HazeTPLA, RainTTPLA, and SnowTTPLA datasets, the content of images is mainly about the transmission
towers and power lines, which are taken from different views with noisy backgrounds, e.g., buildings, plants, roads, and lane lines. In HazeInsPLAD and RainInsPLAD datasets, the images contain 17 unique power line components captured from multiple real-world environmental conditions. In conclusion, the proposed datasets cover the main scenarios of power line autonomous inspection, which can provide strong support for future research about the new PAIR-AW task. (2) We use the average Peak Signal-to-Noise Ratio (PSNR) to measure the information loss of each proposed dataset compared with its clean counterpart. As shown in Fig.~\ref{Fig2} (b), the PSNR of some datasets, such as HazeTTPLA, HazeInsPLAD, are below 20 dB. In addition, the images in all the proposed datasets are aerial images, having special angles, variable target
directions, small-sized objects, and complex backgrounds. The above points suggest that the new PAIR-AW task is very challenging with poor yet complex images to be restored. 
 \begin{figure*}[t]
	%\begin{}
	%\vspace{-0.5cm}
		%\fbox{\rule{0pt}{2in} \rule{.9\linewidth}{0pt}}
		%\includegraphics[scale=0.4]{fig2.jpg}
		\includegraphics[width=1\linewidth]{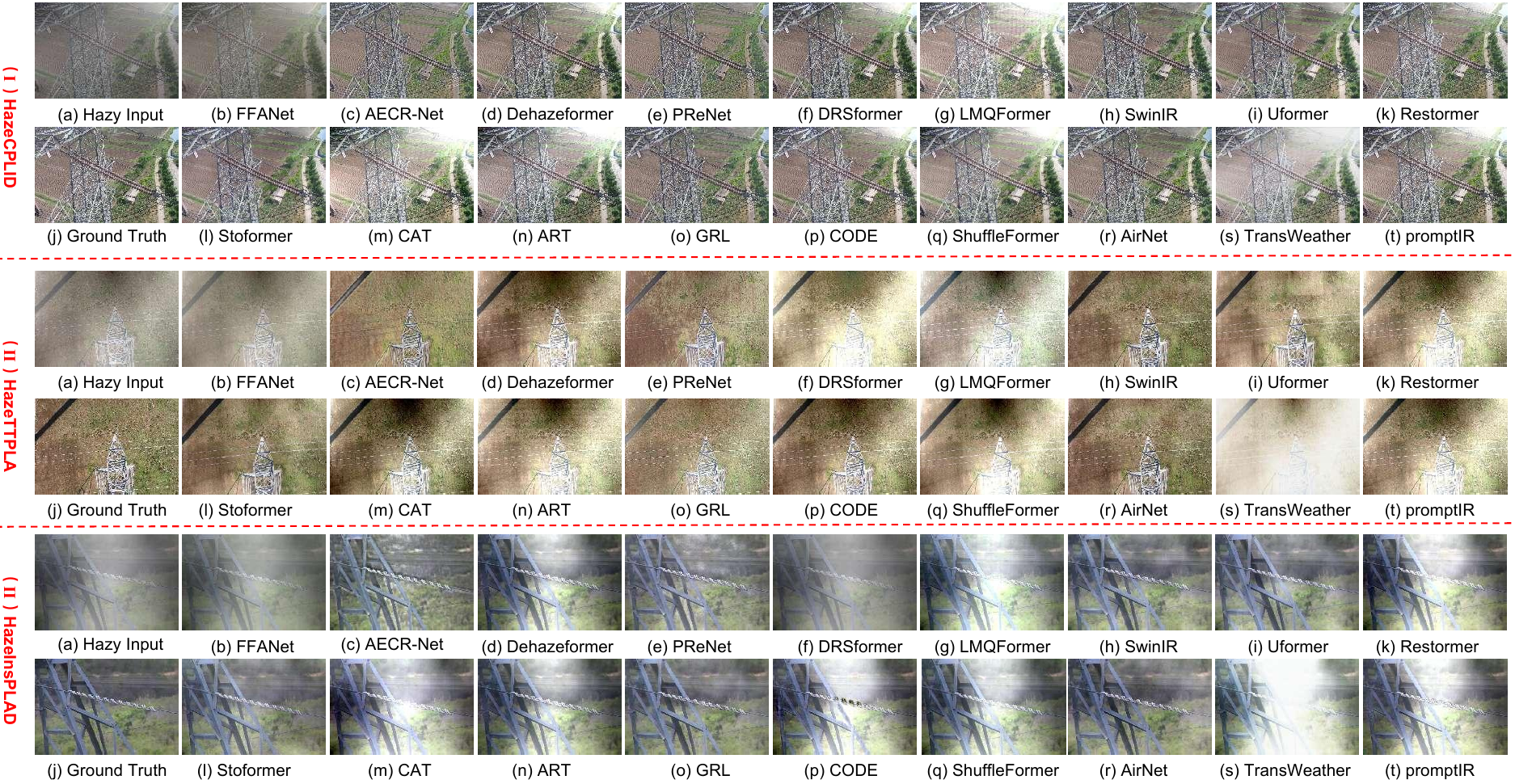}
		%	\vspace*{-0.5	\baselineskip}
		\caption{ Visual comparison results on power line aerial image dehazing task in single-one setting.  Please zoom in on the figure for a better view.}
		\label{Fig4}
	%\end{center}
\end{figure*}

\begin{table*}[ht]
   \renewcommand\tabcolsep{3pt} 
    \footnotesize
   %\addtolength{\tabcolsep}{0.4pt}
    \caption{Quantitative results on power line aerial image dehazing task in single-one setting. The top two results are marked in red and blue, respectively.}
    \label{table3}
    \begin{center}
    \begin{tabular}{p{2.7cm}cccccccccc}
	\Xhline{1.3px}
	% \multirow{2}{*}{Name}，2为所占的行数，此语句可以使得内容垂直居中
	% \multicolumn{2}{c|}{Flag}，2为所占的列数，格式由第二个{}控制
	% \cline{2-3}指本行的2,3列画横线
	\multirow{2}{*}{\textbf{Method}}&\multicolumn{2}{c}{\textbf{HazeCPLID}}&\multicolumn{2}{c}{\textbf{HazeTTPLA}}&\multicolumn{2}{c}{\textbf{HazeInsPLAD}}&\multicolumn{2}{c}{\textbf{Average}}&\multirow{2}{*}{\textbf{\#Param}}&\multirow{2}{*}{\textbf{\#Flops}}\\
\cmidrule(lr){2-3}\cmidrule(lr){4-5}\cmidrule(lr){6-7}\cmidrule(lr){8-9}
 &PSNR$\uparrow$ & SSIM$\uparrow$ &PSNR$\uparrow$ & SSIM$\uparrow$ &PSNR$\uparrow$ & SSIM$\uparrow$&PSNR$\uparrow$ & SSIM$\uparrow$&(M)$\downarrow$ &(G)$\downarrow$\\ 
 %\Xhline{1.2px}
  \hline
FFANet~\cite{qin2020ffa}&20.75&0.8112&19.69&0.7994&19.54&0.8729&19.99&0.8278&4.45&72.08\\
AECR-Net~\cite{wu2021contrastive}&23.35&0.8655&19.87&0.7555&21.75&0.8227&21.65&0.8145&2.59&8.90\\
Dehazeformer~\cite{song2023vision}&28.34&0.9616&26.43&0.9524&\textcolor{red}{28.25}&\textcolor{red}{0.9816}&27.67&0.9652&25.45& 69.24\\
PReNet~\cite{ren2019progressive}&24.01&0.9197&19.71&0.8324&22.56&0.8886&22.09&0.8802&0.17&16.56\\
DRSformer~\cite{chen2023learning}&27.84&\textcolor{blue}{0.9675}&25.72&0.9608&19.48&0.8718&24.34&0.9333&33.65 &55.43\\
LMQFormer~\cite{lin2023lmqformer}&27.40&0.9623&26.53&0.9493&26.20&0.9709&26.71&0.9608&2.18&5.61\\
SwinIR~\cite{liang2021swinir} &26.24&0.9592&24.54&0.9305&26.57&0.9711&25.78&0.9536&7.78&126.51\\
Uformer~\cite{wang2022uformer}&27.38&0.9621&25.33&0.9433&26.84&0.9740&26.51&0.9598&50.88& 89.46\\
%MAXIM~\cite{tu2022maxim}\\
Restormer~\cite{zamir2022restormer}&\textcolor{blue}{27.85}&0.9657&\textcolor{red}{28.06}&\textcolor{blue}{0.9653}&\textcolor{blue}{28.03}&0.9763&\textcolor{red}{27.98}&\textcolor{blue}{0.9691}&28.63& 39.71  \\
CAT~\cite{chen2022cross}&26.85&0.9547&26.93&0.9463&27.90&0.9705&27.22&0.9571&25.77& 33.95\\
Stoformer~\cite{xiao2022stochastic}&27.44&0.9653&25.05&0.9499&22.55&0.9308&25.01&0.9486&50.47&123.97\\
ShuffleFormer~\cite{xiao2023random}&26.69&0.9613&26.85&0.9576&27.81&0.9758&27.11&0.9649&30.75&13.28\\
CODE~\cite{zhao2023comprehensive}&27.18&0.9606&26.59&0.9594&25.47&0.9496&26.41&0.9565&12.23&11.3\\
ART~\cite{2023ART}&27.62&0.9442&25.59&0.8987&27.88&0.9532&27.03&0.9320&25.7&33.71\\
GRL~\cite{li2023efficient}&24.68&0.9406&21.54&0.9089&25.59&0.9632&23.93&0.9375&3.29&51.86\\
AirNet~\cite{li2022all}&26.40&0.9570&25.43&0.9390&27.28&0.9290&26.37&0.9416&7.6&302.3\\
TransWeather~\cite{valanarasu2022transweather}&25.57&0.8934&22.94&0.8284&24.76&0.9436&24.42&0.8884&38.05&1.56\\
promptIR~\cite{potlapalli2023promptir}&\textcolor{red}{28.06}&\textcolor{red}{0.9668}&\textcolor{blue}{27.47}&\textcolor{red}{0.9698}&27.60&\textcolor{blue}{0.9800}&\textcolor{blue}{27.71}&\textcolor{red}{0.9722}&34.12&35.25\\
%\hdashline[2pt/3.5pt]
 
\Xhline{1.3px}
    \end{tabular}
    \end{center}
\end{table*}

\section{Baseline Methods}
Image restoration under adverse weather is a classical low-level computer vision task, which is well-known as the tasks of image dehazing~\cite{he2009single1,2021A,guo2022image,singh2022visibility,2016DehazeNet},  image deraining~\cite{zhang2023data,2020Single} and image desnowing~\cite{2017DesnowNet}. Many brilliant works~\cite{li2017aod,fu2019lightweight,qiu2023mb,dong2020multi,xiao2022image,zhang2021deep} have been proposed to solve these tasks. According to the type number of weather removal, existing methods can be broadly categorized into three groups single adverse weather removal, multiple adverse weather removal, and all-in-one adverse weather removal. We choose the representative methods in each group as the baseline methods for the new task of PLAIR-AW, which will be described  as follows:
\begin{figure*}[t]
	%\begin{}
	%\vspace{-0.5cm}
		%\fbox{\rule{0pt}{2in} \rule{.9\linewidth}{0pt}}
		%\includegraphics[scale=0.4]{fig2.jpg}
		\includegraphics[width=1\linewidth]{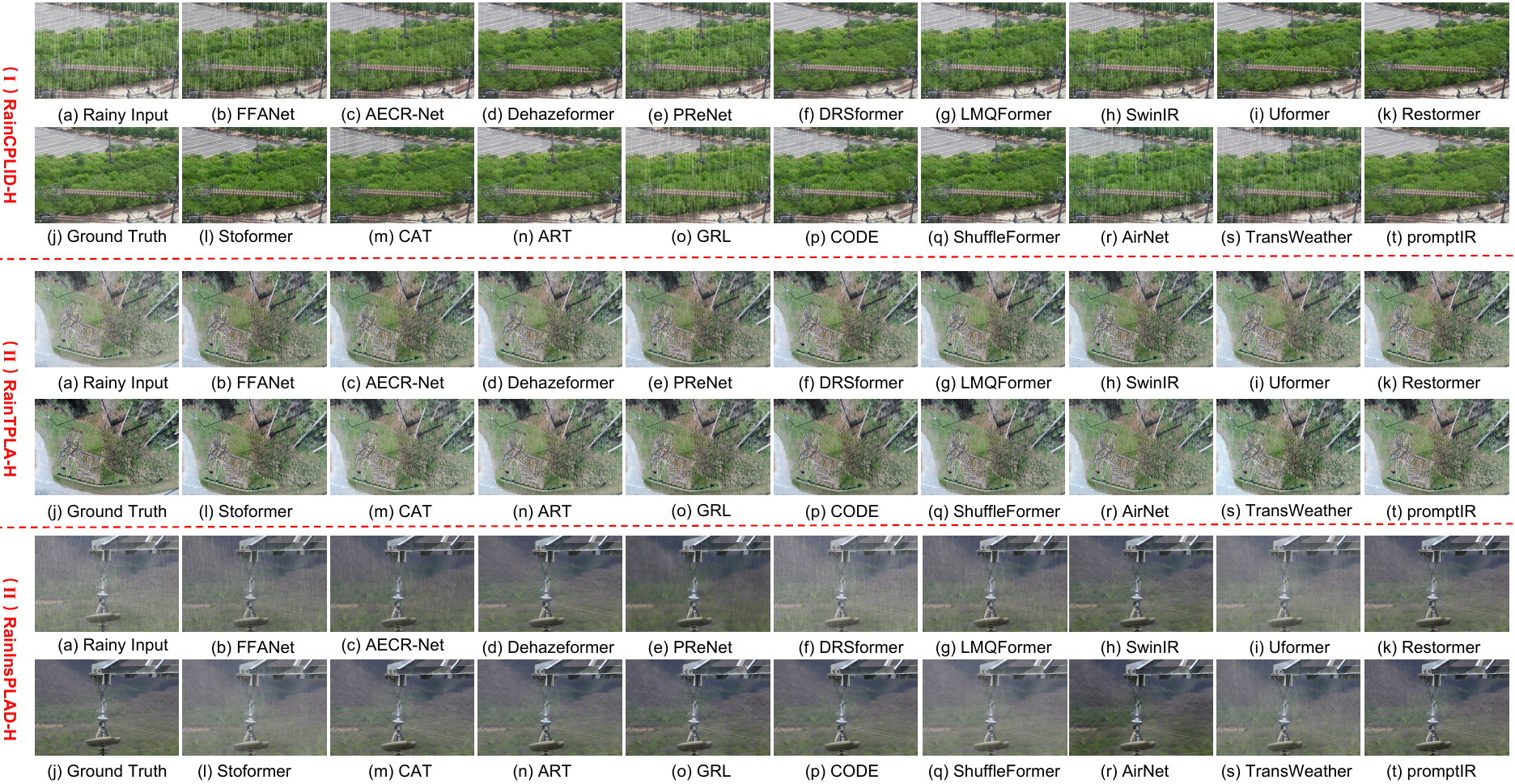}
		%	\vspace*{-0.5	\baselineskip}
		\caption{ Visual comparison results on power line aerial image deraining task in single-one setting. Please zoom in the figure for a better view.}
		\label{Fig5}
	%\end{center}
\end{figure*}
\begin{table*}[ht]
   \renewcommand\tabcolsep{1.7pt} 
     \footnotesize
   %\addtolength{\tabcolsep}{0.4pt}
    \caption{Quantitative results on power line aerial image deraining task in single-one setting. The top two results are marked in red and blue, respectively.}
    \label{Table4}
    \begin{center}
    \begin{tabular}{p{2.2cm}cccccccccccccccc}
	\Xhline{1.3px}
	% \multirow{2}{*}{Name}，2为所占的行数，此语句可以使得内容垂直居中
	% \multicolumn{2}{c|}{Flag}，2为所占的列数，格式由第二个{}控制
	% \cline{2-3}指本行的2,3列画横线
	\multirow{2}{*}{\textbf{Method}}&\multicolumn{2}{c}{\textbf{RainCPLID-L}}&\multicolumn{2}{c}{\textbf{RainCPLID-H}}&\multicolumn{2}{c}{\textbf{RainTTPLA-L}}&\multicolumn{2}{c}{\textbf{RainTTPLA-H}}&\multicolumn{2}{c}{\textbf{RainInsPLAD-L}}&\multicolumn{2}{c}{\textbf{RainInsPLAD-H}}&\multicolumn{2}{c}{\textbf{Average}}&\multirow{2}{*}{\textbf{\#Param}}&\multirow{2}{*}{\textbf{\#Flops}}\\
\cmidrule(lr){2-3}\cmidrule(lr){4-5}\cmidrule(lr){6-7}\cmidrule(lr){8-9}\cmidrule(lr){10-11}\cmidrule(lr){12-13}\cmidrule(lr){14-15}
 &PSNR$\uparrow$ & SSIM$\uparrow$ &PSNR$\uparrow$ & SSIM$\uparrow$ &PSNR$\uparrow$ & SSIM$\uparrow$&PSNR$\uparrow$ & SSIM$\uparrow$&PSNR$\uparrow$ & SSIM$\uparrow$&PSNR$\uparrow$ & SSIM$\uparrow$&PSNR$\uparrow$ & SSIM$\uparrow$&(M)$\downarrow$ &(G)$\downarrow$\\ 
 %\Xhline{1.2px}
  \hline
FFANet~\cite{qin2020ffa}&29.55&0.9281&26.80&0.8779&30.87&0.9604&28.80&0.9318&29.05&0.9314	&27.48&0.9075&28.75&0.9228&4.45&72.08\\
AECR-Net~\cite{wu2021contrastive}&29.22&0.8850&26.92&0.8442&27.42&0.8338&26.32&0.8020&31.87&0.9467	&30.39&0.9295&28.69&0.8735&2.59&8.90\\
Dehazeformer~\cite{song2023vision}&32.61&0.9456&30.73&0.9357&33.70&0.9675&31.24&0.9571&38.01&\textcolor{blue}{0.9820}&36.07&\textcolor{blue}{0.9789}&33.72&0.9611&25.45& 69.24\\
PReNet~\cite{ren2019progressive}&30.97&0.9263&27.95&0.8892&31.10&0.9515&28.99&0.9300&33.77&0.9538	&31.84&0.9380&30.77&0.9314&0.17&16.56\\
DRSformer~\cite{chen2023learning}&34.75&\textcolor{red}{0.9619}&\textcolor{blue}{33.03}&\textcolor{blue}{0.9594}&34.57&0.9728&32.61&0.9698&26.69&0.9227	&25.89&0.8985&31.25&0.9475&33.65 &55.43\\
LMQFormer~\cite{lin2023lmqformer}&32.31&0.9337&29.84&0.9199&32.26&0.9545&29.55&0.9323&35.95&0.9672	&33.64&0.9549&32.25&0.9437&2.18&5.61\\
SwinIR~\cite{liang2021swinir} &30.74&0.9362&28.64&0.9107&31.55&0.9607&28.93&0.9321&35.86&0.9738&33.41&0.9621&31.52&0.9459&7.78&126.51\\
Uformer~\cite{wang2022uformer}&33.45&0.9532&31.93&0.9497&33.13&0.9654&30.91&0.9568&37.57&0.9793	&35.24&0.9746&33.70&0.9630&50.88& 89.46\\
%MAXIM~\cite{tu2022maxim}\\
Restormer~\cite{zamir2022restormer}&\textcolor{blue}{34.80}&0.9616&32.34&0.9574&\textcolor{blue}{34.94}&\textcolor{red}{0.9755}&\textcolor{blue}{32.83}&\textcolor{red}{0.9721}&\textcolor{red}{38.38}&\textcolor{red}{0.9825}&\textcolor{red}{36.44}&\textcolor{red}{0.9800}&\textcolor{red}{34.95}&\textcolor{red}{0.9715}&26.13&35.25  \\
CAT~\cite{chen2022cross}&32.91&0.9428&31.83&0.9433&31.86&0.9584&29.64&0.9440&38.05&0.9802	&35.51&0.9728&33.30&0.9569&15.01&22.03\\
Stoformer~\cite{xiao2022stochastic}&33.39&0.9486&31.00&0.9477&34.42&0.9718&31.74&0.9638&37.77&0.9805&35.38&0.9756&33.28&0.9646&50.47&123.97\\
ShuffleFormer~\cite{xiao2023random}&34.15&0.9516&32.33&0.9539&33.03&0.9648&30.24&0.9517&\textcolor{blue}{38.03}&0.9817&\textcolor{blue}{36.23}&0.9783&34.00&0.9636&30.75&13.28\\
CODE~\cite{zhao2023comprehensive}&32.30&0.9365&31.30&0.9421&31.05&0.9503&29.02&0.9340&37.99&0.9809	&36.22&0.9781&32.98&0.9536&12.23&11.3\\
ART~\cite{2023ART}&32.54&0.9450&31.43&0.9446&32.46&0.9608&30.84&0.9466&37.60&0.9796	&35.86&0.9755&33.45&0.9586&25.7&33.71\\
GRL~\cite{li2023efficient}&30.84&0.9219&27.85&0.8858&31.21&0.9556&28.94&0.9300&35.80&0.9761	&33.82&0.9676&31.41&0.9395&3.29&51.86\\
AirNet~\cite{li2022all}&31.02&0.9343&28.24&0.8962&31.61&0.9628&28.99&0.9434&37.46&0.9787	&35.50&0.9739&32.13&0.9482&7.6&302.3\\
TransWeather~\cite{valanarasu2022transweather}&32.58&0.9298&30.11&0.9205&26.17&0.7621&25.07&0.7243&32.86&0.9357&31.18&0.9242&29.66&0.8661&38.05&1.56\\
promptIR~\cite{potlapalli2023promptir}&\textcolor{red}{34.89}&\textcolor{blue}{0.9605}&\textcolor{red}{33.19}&\textcolor{red}{0.9638}&\textcolor{red}{35.24}&\textcolor{red}{0.9781}&\textcolor{red}{33.18}&\textcolor{blue}{0.9705}&37.45&0.9793&35.06&0.9733&\textcolor{blue}{34.83}&\textcolor{blue}{0.9709}&34.12&35.25\\
%\hdashline[2pt/3.5pt]
 
\Xhline{1.3px}
    \end{tabular}
    \end{center}
\end{table*}
 \begin{figure*}[t]
	%\begin{}
	%\vspace{-0.5cm}
		%\fbox{\rule{0pt}{2in} \rule{.9\linewidth}{0pt}}
		%\includegraphics[scale=0.4]{fig2.jpg}
		\includegraphics[width=1\linewidth]{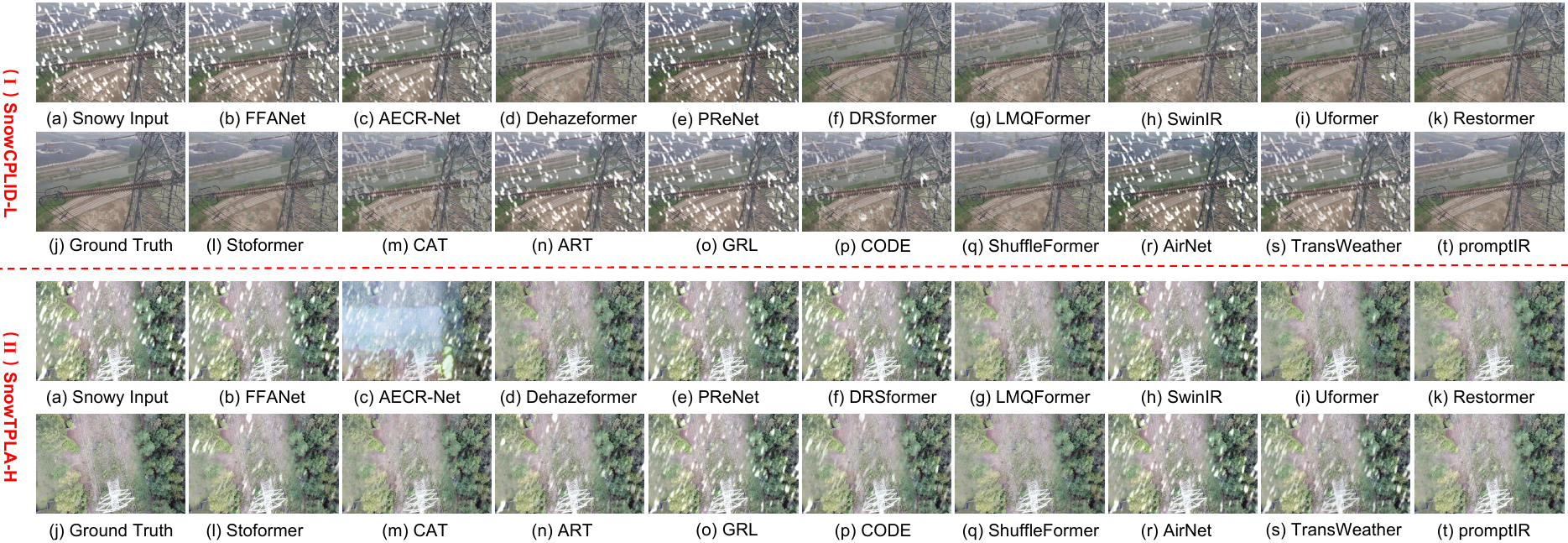}
		%	\vspace*{-0.5	\baselineskip}
		\caption{ Visual comparison results on power line aerial image desnowing task in single-one setting. Please zoom in the figure for a better view.}
		\label{Fig6}
	%\end{center}
\end{figure*}
\begin{table*}[ht]
   \renewcommand\tabcolsep{1.5pt} 
     \footnotesize
   %\addtolength{\tabcolsep}{0.4pt}
    \caption{Quantitative results on power line aerial image desnowing task in the single-one setting. The top two results are marked in red and blue, respectively.}
    \label{Table5}
    \begin{center}
    \begin{tabular}{p{2.2cm}cccccccccccccccc}
	\Xhline{1.3px}
	% \multirow{2}{*}{Name}，2为所占的行数，此语句可以使得内容垂直居中
	% \multicolumn{2}{c|}{Flag}，2为所占的列数，格式由第二个{}控制
	% \cline{2-3}指本行的2,3列画横线
	\multirow{2}{*}{\textbf{Method}}&\multicolumn{2}{c}{\textbf{SnowCPLID-S}}&\multicolumn{2}{c}{\textbf{SnowCPLID-M}}&\multicolumn{2}{c}{\textbf{SnowCPLID-L}}&\multicolumn{2}{c}{\textbf{SnowTTPLA-S}}&\multicolumn{2}{c}{\textbf{SnowTTPLA-M}}&\multicolumn{2}{c}{\textbf{SnowTTPLA-L}}&\multicolumn{2}{c}{\textbf{Average}}&\multirow{2}{*}{\textbf{\#Param}}&\multirow{2}{*}{\textbf{\#Flops}}\\
\cmidrule(lr){2-3}\cmidrule(lr){4-5}\cmidrule(lr){6-7}\cmidrule(lr){8-9}\cmidrule(lr){10-11}\cmidrule(lr){12-13}\cmidrule(lr){14-15}
 &PSNR$\uparrow$ & SSIM$\uparrow$ &PSNR$\uparrow$ & SSIM$\uparrow$ &PSNR$\uparrow$ & SSIM$\uparrow$&PSNR$\uparrow$ & SSIM$\uparrow$&PSNR$\uparrow$ & SSIM$\uparrow$&PSNR$\uparrow$ & SSIM$\uparrow$&PSNR$\uparrow$ & SSIM$\uparrow$&(M)$\downarrow$ &(G)$\downarrow$\\ 
 %\Xhline{1.2px}
  \hline
FFANet~\cite{qin2020ffa}&41.30&0.9695&35.98&0.9235&29.86&0.9187&41.12&0.9754&37.86&0.9687&	27.48&0.9114&35.6&0.9412&4.45& 72.08\\
AECR-Net~\cite{wu2021contrastive}&36.07&0.9558&33.19&0.9230&28.96&0.9033&25.72&0.7046&	25.23&0.6860&22.63&0.6439&28.63&0.8027&2.59&8.90\\
Dehazeformer~\cite{song2023vision}&41.55&0.9753&36.92&0.9442&33.26&0.9450&39.64&0.9800&	37.62&0.9778&31.63&0.9527&36.77&0.9625&25.45& 69.24\\
PReNet~\cite{ren2019progressive}&38.49&0.9604&34.17&0.9153&31.66&0.9315&39.86&0.9762&	37.02&0.9698&27.40&0.9168&34.76&0.9450&0.17&16.56\\
DRSformer~\cite{chen2023learning}&42.70&0.9816&38.68&0.9558&36.67&0.9675&41.05&0.9771&	37.99&0.9696&27.67&0.9184&37.46&0.9616&33.65 &55.43\\
LMQFormer~\cite{lin2023lmqformer}&42.04&0.9749&36.25&0.9286&31.52&0.9294&39.90&0.9773&	37.64&0.9736&30.43&0.9412&36.29&0.9541&2.18&5.61\\
SwinIR~\cite{liang2021swinir}&40.94&0.9722&35.40&0.9279&32.53&0.9383 &39.66&0.9791&	37.54&0.9738&29.53&0.9335&35.93&0.9541&7.78&126.51\\
Uformer~\cite{wang2022uformer}&41.84&0.9765&36.86&0.9405&33.21&0.9484&39.79&0.9805&	38.37&0.9785&32.35&0.9549&37.07&0.9632&13.02&21.89\\
%MAXIM~\cite{tu2022maxim}\\
Restormer~\cite{zamir2022restormer}&\textcolor{blue}{42.99}&\textcolor{red}{0.9834}&\textcolor{blue}{38.68}&\textcolor{blue}{0.9641}&\textcolor{blue}{36.85}&\textcolor{blue}{0.9703}&\textcolor{blue}{41.26}&\textcolor{blue}{0.9831}&	\textcolor{red}{39.38}&\textcolor{blue}{0.9826}&\textcolor{red}{33.82}&\textcolor{red}{0.9681}&\textcolor{blue}{38.82}&\textcolor{blue}{0.9752}&28.63& 39.71  \\
CAT~\cite{chen2022cross}&41.28&0.9725&35.98&0.9326&30.09&0.9140&39.49&0.9798&37.45&0.9776&	30.82&0.9468&35.85&0.9538&25.77 &33.95\\
Stoformer~\cite{xiao2022stochastic}&42.56&0.9816&37.93&0.9584&36.08&0.9644&41.12&0.9768&	37.93&0.9698&27.38&0.9087&37.16&0.9599&50.47&123.97\\
ShuffleFormer~\cite{xiao2023random}&41.67&0.9764&37.08&0.9490&35.09&0.9582&40.60&0.9828	&38.78&0.9813&33.14&0.9643&37.72&0.9686&30.75&13.28\\
CODE~\cite{zhao2023comprehensive}&40.30&0.9702&35.06&0.9234&30.45&0.9163&38.68&0.9766&	36.69&0.9685&29.63&0.9329&35.13&0.9479&12.23&11.3\\
ART~\cite{2023ART}&40.33&0.9705&35.12&0.9205&29.65&0.9180&40.21&0.9786&37.18&0.9732&30.58& 0.9392&35.51&0.9500&25.7&33.71\\
GRL~\cite{li2023efficient}&40.48&0.9696&35.24&0.9237&29.07&0.9080&39.90&0.9797&37.58&0.9747&	28.84&0.9299&35.18&0.9476&3.29&51.86\\
AirNet~\cite{li2022all}&40.20&0.9732&35.25&0.9326&30.42&0.9229&37.55&0.9738&	36.35&0.9716&29.28&0.9362&34.50&0.9517&7.6&302.3\\
TransWeather~\cite{valanarasu2022transweather}&33.52&0.9145&31.37&0.8736&	27.80&0.8546&25.95&0.7213&25.80&0.7094&	23.25&0.6741&27.94&0.7912&38.05&1.56\\
promptIR~\cite{potlapalli2023promptir}&\textcolor{red}{43.39}&\textcolor{blue}{0.9830}&\textcolor{red}{39.48}&\textcolor{red}{0.9691}&\textcolor{red}{37.50}&\textcolor{red}{0.9724}&\textcolor{red}{41.59}&\textcolor{red}{0.9838}&	\textcolor{blue}{39.32}&\textcolor{red}{0.9828}&\textcolor{blue}{33.29}&\textcolor{blue}{0.9673}&\textcolor{red}{39.09}&\textcolor{red}{0.9764}&34.12&35.25\\
%\hdashline[2pt/3.5pt]
 
\Xhline{1.3px}
    \end{tabular}
    \end{center}
\end{table*}
\subsection{Single Adverse Weather Removal}
Single adverse weather removal is referred to as designing a specific method for a certain weather removal task. In this context, we choose the representative methods of FFANet~\cite{qin2020ffa}, AECR-Net~\cite{wu2021contrastive}, Dehazeformer~\cite{song2023vision} for dehazing, PReNet~\cite{ren2019progressive}, DRSformer~\cite{chen2023learning} for deraining, LMQFormer~\cite{lin2023lmqformer} for desnowing, which are presented as following:
\begin{itemize}
\item [$\diamond$] FFANet~\cite{qin2020ffa}: The framework of FFANet is shown in Fig.~\ref{Fig3} (a). The hazy image is passed into a convolutional layer to extract shallow features, which are then fed into N-group architectures. The output features are concatenated to be fused together by the proposed feature attention module. After that, the features are reconstructed to the clean output with the global residual learning connection. 

\item [$\diamond$]AECR-Net~\cite{wu2021contrastive}: As shown in Fig.~\ref{Fig3} (b), it mainly owns autoencoder-like (AE) architecture and constructive learning strategy. Specifically, AE  consists of a downsampling module, six feature attention blocks, a dynamic feature enhancement block, an upsampling module, and two
adaptive mixup operations. Meanwhile, it set the clean image and the hazy image as the positive and negative samples for the degraded output, thereby yielding the contrastive regularization loss.

%\begin{equation}
 %   \label{eq:equ11}
 % \mathscr{L}_{CR}(\hat{I}, I, H)=\sum_{i=1}^N w_i \frac{{\|\phi_i(I(x,y))-\phi_i(\hat{I}(x,y))\|}_1}{{\|\phi_i(H(x,y))-\phi_i(\hat{I}(x,y))\|}_1},
%\end{equation}
%where, $\phi_i$ is the $i$-th hidden feature layer of the pre-trained VGG network, $w_i$
%is a weight coefficient. $I(x,y)$, $H(x,y)$ and $\hat{I}(x,y)$ represent clean image, hazy image, and the output image, respectively.

\item [$\diamond$]Dehazeformer~\cite{song2023vision}: As shown in Fig.~\ref{Fig3} (c), it is arranged into a U-shaped structure with basic Dehzeformer blocks, which are improved upon the popular Swin Transformer~\cite{liu2021swin}. The core improvements mainly include the SK fusion and soft reconstruction layers, which have replaced the concatenation fusion layer and global residual learning.

\item [$\diamond$]PReNet~\cite{ren2019progressive}: As shown in Fig.~\ref{Fig3} (d), it begins with a basic shallow residual network with five residual blocks, which are then developed into multiple stages with recursively unfolding operations. Moreover, a recurrent layer is introduced to exploit the dependencies of deep features across recursive stages.

\item [$\diamond$]DRSformer~\cite{chen2023learning}: As is shown in Fig.~\ref{Fig3} (e), it takes a U-shaped structure with the basic Sparse Transformer Block (STB). The core elements of STB are Top-k sparse attention (TKSA) and Mixed-scale feed-forward network (MSFN). The former explores a learnable top-k selection operator to keep the most useful self-attention values for better feature aggregation, while the latter utilizes the multi-scale depth-wise convolution paths to obtain rich multi-scale representations.

\item [$\diamond$]LMQFormer~\cite{lin2023lmqformer}: As shown in Fig.~\ref{Fig3} (f), it has two paths of Laplace-VQVAE and MQFormer. The first path filters the input image using a Laplace operator and then obtains the coarse mask using a multi-scale encoder-decoder with the Codebook at two low scales. In the second path, the coarse mask and input image are firstly fused and then encoded by two parallel encoders, a hybrid decoder and a modified ConvNeXtBlock. Finally, Pixel Detail Enhancement (PDE) learns further details on the original scale.
\end{itemize}
 \begin{figure*}[t]
	%\begin{}
	%\vspace{-0.5cm}
		%\fbox{\rule{0pt}{2in} \rule{.9\linewidth}{0pt}}
		%\includegraphics[scale=0.4]{fig2.jpg}
		\includegraphics[width=1\linewidth]{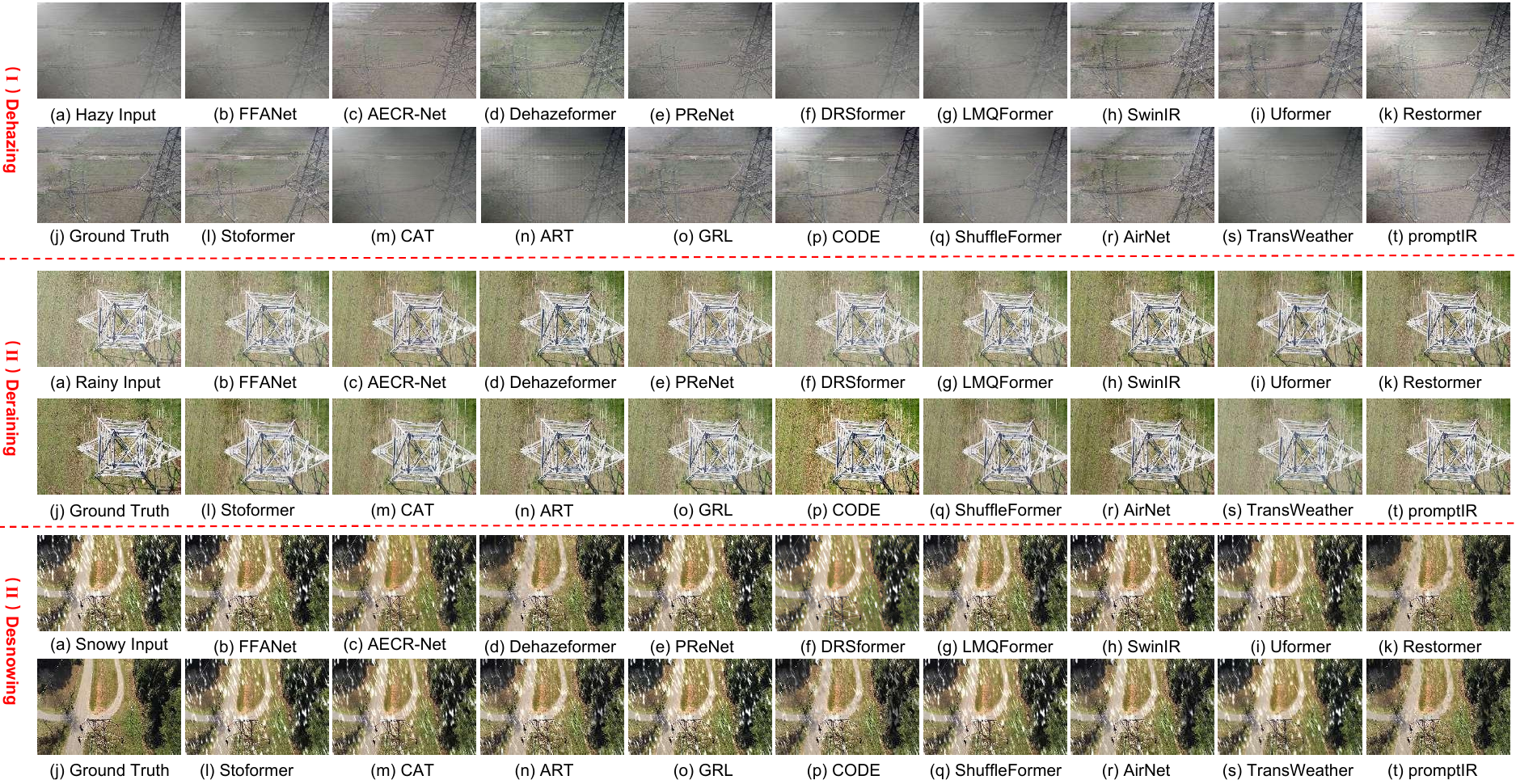}
		%	\vspace*{-0.5	\baselineskip}
		\caption{ Visual comparison results on power line aerial image dehazing, draining, and desnowing tasks in the all-in-one setting. Please zoom in the figure for a better view.}
		\label{Fig7}
	%\end{center}
\end{figure*}
\begin{table*}[ht]
   \renewcommand\tabcolsep{1.7pt} 
    \footnotesize
   %\addtolength{\tabcolsep}{0.4pt}
    \caption{Quantitative results on power line aerial image dehazing task, deraining task, and desnowing task in the all-in-one setting. The top two results are marked in red and blue, respectively.}
    \label{Table6}
    \begin{center}
    \begin{tabular}{p{2.2cm}cccccccccccccccccc}
	\Xhline{1.3px}
	% \multirow{2}{*}{Name}，2为所占的行数，此语句可以使得内容垂直居中
	% \multicolumn{2}{c|}{Flag}，2为所占的列数，格式由第二个{}控制
	% \cline{2-3}指本行的2,3列画横线
	\multirow{3}{*}{\textbf{Method}} & \multicolumn{4}{c}{\textbf{Dehazing}}& \multicolumn{4}{c}{\textbf{Deraining}}&\multicolumn{6}{c}{\textbf{Desnowing}}&\multicolumn{2}{c}{\multirow{2}{*}{\textbf{Average}}}\\
\cmidrule(lr){2-5}\cmidrule(lr){6-9}\cmidrule(lr){10-15}
&\multicolumn{2}{c}{HazeCPLID} & \multicolumn{2}{c}{HazeTTPLA}  &\multicolumn{2}{c}{RainCPLID-L} & \multicolumn{2}{c}{RainCPLID-H}&\multicolumn{2}{c}{SnowTTPLA-S} & \multicolumn{2}{c}{SnowTTPLA-M}&\multicolumn{2}{c}{SnowTTPLA-L}   \\  
\cmidrule(lr){2-3}\cmidrule(lr){4-5}\cmidrule(lr){6-7}\cmidrule(lr){8-9}\cmidrule(lr){10-11}\cmidrule(lr){12-13}\cmidrule(lr){14-15}\cmidrule(lr){16-17}
 &PSNR$\uparrow$ & SSIM$\uparrow$ &PSNR$\uparrow$ & SSIM$\uparrow$&PSNR$\uparrow$ & SSIM$\uparrow$&PSNR$\uparrow$ & SSIM$\uparrow$&PSNR$\uparrow$ & SSIM$\uparrow$ &PSNR$\uparrow$ & SSIM$\uparrow$ &PSNR$\uparrow$ & SSIM$\uparrow$&PSNR$\uparrow$ & SSIM$\uparrow$ \\ 
 %\Xhline{1.2px}
  \hline
FFANet~\cite{qin2020ffa}&20.49&0.8274&18.83&0.7941&25.39&0.9276&25.46&0.9007&37.61&0.9706&36.33&0.9670&27.43&0.9160&27.36&0.9004\\
AECR-Net~\cite{wu2021contrastive}&21.69&0.8179&19.68&0.7837&26.28&0.8553&25.92&0.8367&24.62&0.7479&24.36&0.7384&22.30&0.6989&23.55&0.7971\\
Dehazeformer~\cite{song2023vision}&26.00&0.9611&\textcolor{red}{23.35}&\textcolor{red}{0.9133}&\textcolor{blue}{31.61}&0.9541&30.23&0.9393&34.67&0.9632&33.99&0.9648&28.75&0.9248&29.8&\textcolor{blue}{0.9458}\\
PReNet~\cite{ren2019progressive}&20.86& 0.8567&19.05&0.8122&26.05&0.9018&26.10& 0.8770&34.83& 0.9576&34.47&0.9636&26.99&0.9181&26.90&0.8981\\
DRSformer~\cite{chen2023learning}&20.89&0.8218&19.01&0.7919&23.74&0.9231&23.60&0.8965&\textcolor{red}{35.36}&0.9607&\textcolor{blue}{34.87}&0.9518&27.86&0.9221&26.47&0.8954\\
LMQFormer~\cite{lin2023lmqformer}&25.67& 0.9519&21.74&0.8813&29.81&0.9316&28.01& 0.9178&34.52&0.9487&33.84& 0.9550&27.78&0.9265&28.76&0.9304\\
SwinIR~\cite{liang2021swinir} &26.11&0.9527&20.21&0.8520&28.38&0.9307&27.19&0.9031&29.59&0.9275&29.42&0.9308&24.36&0.8895&26.45&0.8649\\
Uformer~\cite{wang2022uformer}&26.04&0.9514&20.23&0.8575&27.64&0.9295&26.46&0.9088&\textcolor{blue}{35.26}&0.9637&\textcolor{red}{34.91}&0.9653&27.21&0.9214&28.25&0.9282\\
%MAXIM~\cite{tu2022maxim}\\
Restormer~\cite{zamir2022restormer}&\textcolor{red}{27.25}&\textcolor{red}{0.9641}&21.64&0.8460&31.36&\textcolor{blue}{0.9584}&\textcolor{blue}{30.41}&\textcolor{blue}{0.9499}&33.55&0.9648&34.86&\textcolor{red}{0.9720}&\textcolor{blue}{29.90}&\textcolor{red}{0.9433}&\textcolor{blue}{29.85}&0.9426 \\
CAT~\cite{chen2022cross}&26.33&0.9537&22.19&0.8807&31.30&0.9418&29.45&0.9312&33.23&\textcolor{blue}{0.9659}&33.21&0.9628&27.97&0.9236&29.09&0.9371\\
Stoformer~\cite{xiao2022stochastic}&25.44&0.9555&20.92&\textcolor{blue}{0.9010}&29.67&0.9268&26.99&0.8844&31.33&0.9538&31.34&0.9546&25.39&0.8996&27.29&0.9251\\
ShuffleFormer~\cite{xiao2023random}&26.11&0.9633&21.92&0.9032&30.25&0.9391&29.38&0.9180&33.49&0.9434&33.41&0.9498&26.86&0.9101&28.77&0.9324\\
CODE~\cite{zhao2023comprehensive}&25.27&0.9529&19.84&0.8012&29.38&0.9356&28.90&0.9323&32.34&0.9557&32.03&0.9540&27.81&0.9272&27.93&0.9227\\
ART~\cite{2023ART}&26.32&0.9302&\textcolor{blue}{22.33}&0.8557&31.22&0.9360&29.61&0.9222&33.63&0.9625&33.33&0.9591&27.95&0.9249&29.19&0.9272\\
GRL~\cite{li2023efficient}&24.15&0.9252&19.24&0.8024&27.69&0.9137&27.39&0.8990&31.43&0.9493&31.02&0.9463&25.48&0.9009&26.62&0.9052\\

AirNet~\cite{li2022all}&26.09&0.9510&20.20&0.8830&30.26&0.9445&28.90&0.9270&33.98&0.9647&33.42&0.9624&28.30&0.9300&28.73&0.9375\\
TransWeather~\cite{valanarasu2022transweather}&23.13&0.8755&20.93&0.8334&27.56&0.8700&26.53&0.8572&24.00&0.7524&23.80&0.7503&22.15&0.7050&24.01&0.8062\\
promptIR~\cite{potlapalli2023promptir}&\textcolor{blue}{26.56}&\textcolor{blue}{ 0.9637}&21.71& 0.8681&\textcolor{red}{33.44}& \textcolor{red}{0.9709}&\textcolor{red}{31.97}&\textcolor{red}{0.9654}&33.92&\textcolor{red}{0.9686}&33.40&\textcolor{blue}{ 0.9704}&\textcolor{red}{29.91}&\textcolor{blue}{0.9428}&\textcolor{red}{30.13}&\textcolor{red}{0.9499}\\
%\hdashline[2pt/3.5pt]
 
\Xhline{1.3px}
    \end{tabular}
    \end{center}
\end{table*}

\subsection{Multiple Adverse Weather Removal}
Multiple adverse weather removal is referred to as developing general models customizing for multiple weather removal tasks~\cite{tu2022maxim,zhou2023fourmer}, which has emerged as a hot spot in the image restoration community with the following representatives:
\begin{itemize}
\item [$\diamond$] SwinIR~\cite{liang2021swinir}: As shown in Fig.~\ref{Fig3} (g), it consists of
three modules of (1) shallow feature extraction, (2) deep feature extraction, and (3) high-quality image reconstruction. Specifically, the first module is a 3 × 3 convolutional layer. The second module is composed of residual Swin Transformer Blocks, where Swin transformer layers are assembled together with a residual connection. The third module aggregates the shallow and deep features together, which can help the second module focus on high-frequency information. 

\item [$\diamond$] Uformer~\cite{wang2022uformer}: As shown in Fig.~\ref{Fig3} (h), it mainly arranges the Locally-enhanced Window (LeWin) transformer blocks in a U-shaped structure. LeWin performs non-overlapping window-based self-attention instead of global self-attention. Meanwhile, a learnable multi-scale restoration modulator is added to restore more details.

\item [$\diamond$] Restormer~\cite{zamir2022restormer}: As shown in Fig.~\ref{Fig3} (i), it is generally designed in a multi-scale hierarchical. The core components of the basic block are multi-Dconv head transposed attention (MDTA) and Gated-Dconv feed-forward network (GDFN). MDTA models global context by performing channels-wised attention rather the spatial-based attention, GDFN introduces the gating mechanism for allowing only the useful information to pass further through the network hierarchy. 

\item [$\diamond$]CAT~\cite{chen2022cross}: As shown in Fig.~\ref{Fig3} (j), it consists of three modules of (1) shallow feature extraction, (2) deep feature extraction, and (3) reconstruction. The core component is the second module, which applies a residual connection with the elements of several residual groups and one convolution layer. Each residual group consists of several cross-aggregation transformer blocks and a convolution
layer.

\item [$\diamond$] Stoformer~\cite{xiao2022stochastic}: As shown in Fig.~\ref{Fig3} (k), it utilizes a UNet-style network architecture with StoBlock in the encoder and decoder stage. The basic transformer layer in StoBlock is a variant of window-based self-attention with a stochastic window strategy. 

\item [$\diamond$] ShuffleFormer~\cite{xiao2023random}: As shown in Fig.~\ref{Fig3} (l), it is established with U-shape architecture using the basic block of ShufflleBlock. The basic transformer layer in ShuffleBlock enhanced the non-local interactions of the local window transformer by randomly shuffling the input.

\item [$\diamond$] CODE~\cite{zhao2023comprehensive}: As shown in Fig.~\ref{Fig3}(m), it arranges the basic transformer blocks in a U-shape structure, and each basic block consists of the Condensed Attention (CA) block and the Dual Adaptive (DA) block. CA sequentially performs feature aggregation, attention computation, and feature recovery to efficiently capture the superpixel-wise global feature, while DA takes a dual-way structure in a dynamic weighting fashion to distribute the superpixel-wise globality into each pixel.

\item [$\diamond$] ART~\cite{2023ART}: As shown in Fig.~\ref{Fig3} (n), it takes the asymmetric auto-encoder architecture, which firstly extracts shallow features with a $3 \times 3$ convolutional layer, then extracts deep features with several residual groups, and finally refine the restoration results with a $3 \times 3$ convolutional layer. The core element of each residual group is two successive attention blocks of Dense Attention Block (DAB) and Sparse Attention Block (SAB).

\item [$\diamond$] GRL~\cite{li2023efficient}: As shown in Fig.~\ref{Fig3} (o), it mainly contains (1) feature extraction, (2) representation learning, and (3) image reconstruction. The second part is the backbone network, which takes residual connection with six transformer stage and a convolutional layer. The transformer layer implements a parallel computation of the anchored stripe self-attention, window self-attention, and channel-attention enhanced convolution.
\end{itemize}
\subsection{All-in-one Adverse Weather Removal}
All-in-one adverse weather removal is referred to as developing a unified architecture to be capable of dealing with multiple adverse weather removal tasks simultaneously~\cite{chen2021pre,chen2022learning}, which has developed to be a new research trend in the image restoration community with the following representatives.
\begin{itemize}
\item [$\diamond$]AirNet~\cite{li2022all}: As shown in Fig.~\ref{Fig3} (p), it mainly contains Contrastive-Based Degraded Encoder (CBDE) and Degradation-Guided Restoration Network (DGRN). The core component in CBDE is the supervised contrastive learning to discriminate degradation types, which can efficiently guide image restoration in DGRN. 

\item [$\diamond$]TransWeather~\cite{valanarasu2022transweather}: As shown in Fig.~\ref{Fig3} (q), the framework of TransWeather generally follows auto-encoder architecture. The encoder has intra-patch transformer blocks to extract features from smaller sub-patches created from the main patch. The transformer decoder has learnable weather-type queries to obtain the task features. Then, hierarchical features and task features are forwarded to a convolutional projection block to obtain a clean image.

\item [$\diamond$] PromptIR~\cite{potlapalli2023promptir}: As shown in Fig.~\ref{Fig3} (r), it is designed upon Restormer~\cite{zamir2022restormer} with a new plug-and-play prompt module. The new module encodes degradation-specific information with a set of tunable parameters, which can dynamically guide the decoder to restore the image with various degradation types.
\end{itemize}
\textbf{Remarks and Potentials:} The baseline methods are summarized in Table~\ref{table2}.  All the baseline methods are extensively compared in both single-one and all-in-one settings in our work. We calculate the average results over all the datasets, tasks as well as settings in PSNR. From the result shown in Fig.2 (c), we can see Restormer~\cite{zamir2022restormer} and PromptIR~\cite{potlapalli2023promptir} comprehensively rank the first place and the second place, respectively. The triumph of these two methods is mainly attributed to the powerful long-range dependency modeling capability of the transformers. Yet, the transformers suffer heavy computational and parameter capacity burdens due to their multi-head self-attention, which can not meet the resource-limited application of power line autonomous inspection. Future work is needed to focus on developing new specific restoration models targeting power line autonomous inspection tasks.

\begin{table*}[ht]
   \renewcommand\tabcolsep{2pt} 
    \footnotesize
   %\addtolength{\tabcolsep}{0.4pt}
    \caption{Segmentation performance comparison among the baseline methods, normal condition as well as the method without PAIR-AW (w/o PAIR-AW). The top two results achieved by the baseline methods are marked with red and blue, respectively. The result in normal condition is marked in bold. The result of w/o PAIR-AW is marked with an underline. }
    \label{table7}
    \begin{center}
    \begin{tabular}{p{2.4cm}ccccccccccccccccccc}
	\Xhline{1.3px}
	% \multirow{2}{*}{Name}，2为所占的行数，此语句可以使得内容垂直居中
	% \multicolumn{2}{c|}{Flag}，2为所占的列数，格式由第二个{}控制
	% \cline{2-3}指本行的2,3列画横线
	\multirow{2}{*}{\textbf{Method}}&\multicolumn{5}{c}{\textbf{HazeTTPLA}}&\multicolumn{5}{c}{\textbf{RainTTPLA-H}}&\multicolumn{5}{c}{\textbf{SnowTTPLA-L}}\\
\cmidrule(lr){2-6}\cmidrule(lr){7-11}\cmidrule(lr){12-16}
 &$AP_b^{50}$ & $AP_m^{50}$ &$AP_b^{75}$ & $AP_m^{75}$&$AP_b^{avg}$ &$AP_b^{50}$ & $AP_m^{50}$ &$AP_b^{75}$ & $AP_m^{75}$&$AP^{avg}$  &$AP_b^{50}$ & $AP_m^{50}$ &$AP_b^{75}$ & $AP_m^{75}$&$AP^{avg}$ \\
 %\Xhline{1.2px}
  \hline
Normal&\textbf{57.63}&\textbf{42.26}&\textbf{33.78}&\textbf{21.74}&\textbf{38.85}&\textbf{57.63}&\textbf{42.26}&\textbf{33.78}&\textbf{21.74}&\textbf{38.85}&\textbf{57.63}&\textbf{42.26}&\textbf{33.78}&\textbf{21.74}&\textbf{38.85}\\
w/o PAIR-AW&\uline{53.98}&\uline{33.56}&\uline{30.47}&\uline{18.14}&\uline{34.03}&\uline{53.72}&\uline{37.42}&\uline{30.35}&\uline{19.60}&\uline{35.27}&\uline{53.95}&\uline{37.07}&\uline{28.95}&\uline{19.62}&\uline{34.89}\\
FFANet~\cite{qin2020ffa}&52.50&35.04&29.47&18.27&33.82&55.10&41.45&30.90&21.11&37.14&54.47&36.89&28.47&19.18&34.75\\
AECR-Net~\cite{wu2021contrastive}&42.53&29.77&19.40&13.62&26.33&45.23&31.37&20.74&13.69&27.75&44.57&28.04&23.73&11.48&26.95\\
Dehazeformer~\cite{song2023vision}&56.14&\textcolor{red}{42.73}&31.34&20.31&37.63&\textcolor{red}{55.91}&41.75&\textcolor{red}{32.31}&\textcolor{red}{21.26}&\textcolor{red}{37.80}&\textcolor{blue}{56.41}&\textcolor{blue}{41.45}&29.23&\textcolor{blue}{20.06}&36.78\\
PReNet~\cite{ren2019progressive}&52.76&41.60&25.65&17.83&34.46&55.35&41.49&30.90&20.66&37.10&54.12&37.26&28.04&18.58&34.50\\
DRSformer~\cite{chen2023learning}&50.25&35.50&25.35&14.70&31.45&55.29&41.09&31.27&20.52&37.04&54.03&37.40&28.54&18.99&34.74\\
LMQFormer~\cite{lin2023lmqformer}&50.71&35.53&26.78&17.53&32.63&54.16&41.09&31.34&20.72&36.82&54.72&39.13&28.90&19.72&35.61\\
SwinIR~\cite{liang2021swinir} &\textcolor{red}{57.25}&40.87&\textcolor{blue}{33.33}&20.79&\textcolor{blue}{38.06}&54.98&41.39&31.44&20.36&37.04&55.68&37.54&29.54&19.05&35.45\\
Uformer~\cite{wang2022uformer}&56.24&\textcolor{blue}{42.05}&\textcolor{red}{34.09}&20.56&\textcolor{red}{38.23}&54.44&41.36&31.02&20.64&36.86&55.65&40.66&\textcolor{red}{31.50}&19.84&\textcolor{blue}{36.91}\\
%MAXIM~\cite{tu2022maxim}\\
Restormer~\cite{zamir2022restormer} &55.90&40.55&30.85&18.75&36.51&55.27&41.65&31.44&20.78&37.28&55.70&40.93&\textcolor{blue}{30.99}&18.88&36.62\\
CAT~\cite{chen2022cross}&55.69&40.20&32.96&20.41&37.31&54.52&\textcolor{blue}{41.73}&\textcolor{blue}{31.98}&20.92&37.28&55.30&35.47&28.10&19.30&34.54\\
Stoformer~\cite{xiao2022stochastic}&56.49&42.00&32.55&20.73&37.94&\textcolor{blue}{55.78}&\textcolor{red}{42.27}&31.43&20.88&\textcolor{blue}{37.59}&56.08&40.24&29.07&\textcolor{red}{20.81}&36.55\\
ShuffleFormer~\cite{xiao2023random}&53.25&39.05&28.97&18.04&34.82&55.09&41.46&31.35&\textcolor{blue}{21.23}&37.25&55.92&41.26&30.28&19.50&36.74\\
CODE~\cite{zhao2023comprehensive}&\textcolor{blue}{57.03}&41.09&31.10&\textcolor{red}{21.91}&37.78&55.04&41.53&31.83&20.84&37.31&55.54&38.11&29.55&19.87&35.76\\
ART~\cite{2023ART}&53.86&40.10&32.96&20.41&36.83&55.01&41.44&31.97&21.14&37.39&55.11&39.42&29.66&19.91&36.02\\
GRL~\cite{li2023efficient}&55.96&37.48&30.69&\textcolor{blue}{20.96}&36.27&54.39&41.03&31.24&20.35&36.87&55.60&37.12&30.64&19.35&35.67\\

AirNet~\cite{li2022all}&54.70&36.03&29.08&19.71&34.88&54.29&41.36&30.87&20.62&36.78&54.57&38.27&28.16&19.47&35.09\\
TransWeather~\cite{valanarasu2022transweather}&29.28&20.13&11.76&7.79&17.24&31.38&21.92&13.53&9.11&18.98&44.71&30.53&20.74&14.37&27.58\\
promptIR~\cite{potlapalli2023promptir}&54.80&38.64&30.40&19.53&35.84&55.51&41.51&30.95&20.46&37.10&\textcolor{red}{56.94}&\textcolor{red}{41.85}&30.93&19.75&\textcolor{red}{37.36}\\
%\hdashline[2pt/3.5pt]
 
\Xhline{1.3px}
    \end{tabular}
    \end{center}
\end{table*}
\section{Experiments}
\subsection{Restoration Evaluation}
\subsubsection{Implementation Details}
All the baseline methods are extensively compared in both single-one and all-in-one settings. The single-one setting trains separate models with training datasets proposed in power line aerial image dehazing task, draining, and desnowing tasks. The all-in-one setting trains a unified
model by combining partial training datasets in each restoration task, i.e, HazeCPLID and HazeTTPLA in the power line aerial image dehazing task,
RainCPLID-L, RainCPLID-H in power line aerial image deraining task, and SnowTTPLA-S,
SnowTTPLA-M and SnowTTPLA-L in power line aerial image desnowing task.  For all the datasets,
Adam with a momentum of $\beta_1=0.9,\beta_2=0.999$ and a weight decay of 5e-4
is employed as the optimizer. The maximal epoch number is set to be 200 with the batch
size of 8. Following~\cite{zheng2023curricular,liu2022griddehazenet+,guo2023scanet,cao2021two,zhu2021multi,cheng2023snow}, we select popular metrics of Peak Signal-to-Noise Ratio (PSNR)
and Structure Similarity (SSIM) as the quantitative
measures. Higher values of these metrics indicate better
performance of the methods. Partial trained models have been uploaded to Cloud Drive.\footnote{\textcolor[rgb]{1,0.1,0.5}{https://pan.ntu.edu.cn/l/s1RL7R}}
\subsubsection{Results}
In a single-one setting, the quantitative comparison results on power line aerial image dehazing, deraining, and desnowing tasks are reported in Table~\ref{table3}, Table~\ref{Table4}, and Table~\ref{Table5}, respectively. From the results, we can see that: (1) On the whole, the transformer-based methods like Restormer~\cite{zamir2022restormer}, CAT~\cite{chen2022cross}, ART~\cite{2023ART}, promptIR~\cite{potlapalli2023promptir} perform better than the CNNs-based methods like FFANet~\cite{qin2020ffa}, AECR-Net~\cite{wu2021contrastive} on power line aerial image dehazing, deraining and desnowing tasks in single-one setting. (2) From the average results in Table~\ref{table3}, Table~\ref{Table4}, and Table~\ref{Table5}, we can see that  Restormer~\cite{zamir2022restormer} and promptIR~\cite{potlapalli2023promptir} comprehensively achieve the top two performance in both PSNR and SSIM. We further show the visualization comparison results of power line aerial image dehazing, deraining, and desnowing tasks in Fig.~\ref{Fig4}, Fig.~\ref{Fig5}, Fig.~\ref{Fig6}, respectively. The visualization results appear to be consistent with the above quantitative results, where Restormer~\cite{zamir2022restormer} and promptIR~\cite{potlapalli2023promptir} can generate visually pleasing results with better structures and details. In contrast, some CNNs-based methods like FFANet~\cite{qin2020ffa}, AECR-Net~\cite{wu2021contrastive} still have a faint mist in the restored images as shown in Fig.~\ref{Fig4}. As shown in Fig.~\ref{Fig5}. there still exists rain streak residuals in the restored images of FFANet~\cite{qin2020ffa}, AECR-Net~\cite{wu2021contrastive}. As shown in Fig.~\ref{Fig6}, FFANet~\cite{qin2020ffa}, AECR-Net~\cite{wu2021contrastive} fail to recover well with some snowflake residuals.

In the all-in-one setting, quantitative comparison results are reported in Table~\ref{Table6}. From the results, it can be observed that: (1) Compared with the results in single-one setting shown in Table~\ref{table3}, Table~\ref{Table4}, and Table~\ref{Table5}, the performance of all the methods has significantly reduced, illustrating the all-in-one experimental setting is more challenging than the sing-one setting. (2) Comprehensively speaking,  Restormer~\cite{zamir2022restormer} and promptIR~\cite{potlapalli2023promptir} rank the best and second place in both PSNR and SSIM. Their success is mainly attributed to the advanced transformer blocks, which implement spatial-wise self-attention operations. We further show the visualization comparison results of power line aerial image dehazing, deraining, and desnowing tasks in an all-in-one setting in Fig.~\ref{Fig7}. The visualization results appear to be consistent with the above quantitative results, where Restormer~\cite{zamir2022restormer} and promptIR~\cite{potlapalli2023promptir} can generate visually pleasing results with better structures and details. In contrast, some methods like FFANet~\cite{qin2020ffa}, AECR-Net~\cite{wu2021contrastive} fail to recover well with distorted results.

\subsection{Autonomous Inspection-based Evaluation}
To investigate whether the Power Line Aerial Image Restoration under Adverse Weather (PLAIR-AW) task benefits
power line autonomous inspection, we apply the state-of-the-art real-time instance segmentation model of YOLACT~\cite{bolya2019yolact} to evaluate the images after PAIR-AW task. This experiment is implemented on HazeTTPLA, RainTTPLA-H, and SnowTTPLA-L. Following~\cite{he2023fastinst}, we calculate the standard box Average Precision (AP) under different IoU thresholds as the evaluation metric. The average precision is calculated for both bounding boxes and instance mask, which are denoted as $AP_b$ and $AP_m$, respectively.  The precision scores are evaluated with two cases, i.e. AP with the overlap value of 50\% and 75\%, resulting in $AP_b^{50\%}$, $AP_b^{75\%}$, $AP_m^{50\%}$, $AP_m^{75\%}$. We also calculate the average result over $AP_b^{50\%}$, $AP_b^{75\%}$, $AP_m^{50\%}$, $AP_m^{75\%}$ to comprehensively compare the methods, which is denoted as $AP^{avg}$. The segmentation results of each baseline method are shown in Table~\ref{table7}. 

From the results, we have the following observation: (1) On the power line aerial image dehazing task, compared with the method of w/o PAIR-AW, FFANet~\cite{qin2020ffa}, AECR-Net~\cite{wu2021contrastive}, DRSformer~\cite{chen2023learning}, LMQFormer~\cite{lin2023lmqformer}, TransWeather~\cite{valanarasu2022transweather} do not obtain improvements. On the power line aerial image deraining task, compared with the method of w/o PAIR-AW, AECR-Net~\cite{wu2021contrastive}, TransWeather~\cite{valanarasu2022transweather} do not obtain improvements. On the power line aerial image desnowing task, compared with the method of w/o PAIR-AW, FFANet~\cite{qin2020ffa}, AECR-Net~\cite{wu2021contrastive}, PReNet~\cite{ren2019progressive}, DRSformer~\cite{chen2023learning}, CAT~\cite{chen2022cross}, TransWeather~\cite{valanarasu2022transweather} do not obtain improvements. The above analyses suggest these methods fail in the autonomous inspection-based evaluation. (2)~The best performance is achieved by Uformer~\cite{wang2022uformer}, Dehazeformer~\cite{song2023vision}, promptIR~\cite{potlapalli2023promptir} on dehazning task, deraning task, and desnowing task, respectively. These methods significantly outperform w/o PAIR-AW by a large gain, suggesting the proposed new task can benefit the power line autonomous inspection under adverse weather. (3) The baseline methods with the best performance are still inferior to the results in normal conditions, demonstrating more advanced image restoration models need to be designed in future work. 

\section{Conclusions and Future Work}
Under adverse weather conditions, the overhead power lines are more susceptible to malfunctions, thereby strengthening inspection is crucial to ensure the stable operation of the power system. In this context, autonomous inspection is greatly superior to manual inspection in the aspects of safety and efficiency. However, our investigation in this paper has verified that aerial images captured in adverse weather are detrimental to modern autonomous inspection methods based on deep learning. To address this problem, we propose a new task of Power Line Aerial Image Restoration under Adverse Weather (PLAIR-AW) to enhance the visible quality of aerial images. Meanwhile, we formulate the general solution pipeline based on deep learning for this new task. Further, to realize the solution, we construct numerous synthetic datasets following reasonable mathematical models. These datasets include HazeCPLID, HazeTTPLA, HazeInsPLAD for power line aerial image dehazing task, RainCPLID, RainTTPLA,
RainInsPLAD for the power line aerial image deraining task, and SnowCPLID, SnowTTPLA for the power line aerial image desnowing task. Moreover, we also provide numerous excellent baseline methods for the new task. These baseline methods have been extensively evaluated on the proposed datasets in both single-one and all-in-one settings.

Since PLAIR-AW is a new task, much work is needed to be carried out in the future. For example, (1) The proposed datasets are artificially synthesized, having bias from the real-world degraded aerial images under adverse weather, thereby producing the domain shift problem in the realistic evaluation. Thus, future work will collect sufficient real-world degraded aerial images, which can assist the existing proposed synthesized datasets to develop the semi-supervised learning strategy. (2) Despite the promising performance of baselines in PLAIR-AW tasks, they still exist the following two issues. On the one hand, they are not customized for aerial images, which have characteristics of special angles, variable target directions, small-sized objects, and complex backgrounds. Future works will concentrate on enhancing the representation of the baseline methods to deal with the hard aerial image restoration. On the other hand, they are not lightweight enough to satisfy the resource-limited demands of UAVs. Thus, future work will focus on model compression techniques to reduce the parameter capacity of restoration models. 
\normalem
\bibliography{IEEE2}
\bibliographystyle{IEEEtran}

\end{document}